\documentclass{article}

\usepackage{microtype}
\usepackage{graphicx}
\usepackage{booktabs}
\usepackage{multirow}
\usepackage{array}
\usepackage{xcolor}
\usepackage{tikz}
\usepackage{alltt}
\newsavebox{\codebox}
\usepackage{hyperref}

\usepackage[preprint]{icml2026}

\usepackage{amsmath}
\usepackage{amssymb}
\usepackage{mathtools}
\usepackage{amsthm}

\newcommand{\sys}{TADA}

\providecommand{\Description}[2][]{}

\newcommand{\suppTopkBootstrap}{Appendix~\ref{app:topk-bootstrap}}
\newcommand{\suppEECost}{Appendix~\ref{app:e2e-cost}}
\newcommand{\suppRQControlled}{Appendix~\ref{app:rq2-controlled}}

\theoremstyle{plain}
\newtheorem{theorem}{Theorem}

\theoremstyle{definition}

\makeatletter
\renewcommand{\printAffiliationsAndNotice}[1]{\global\icml@noticeprintedtrue%
  {\let\thefootnote\relax\footnotetext{\raggedright\hspace*{-\footnotesep}#1%
    Correspondence to: \icmlcorrespondingauthor@text.\ \\ \Notice@String}}}
\makeatother

\icmltitlerunning{Semantic Atoms for Scalable Text Analysis in Data Analytics}

\begin{document}
\raggedbottom

\twocolumn[
  \icmltitle{Semantic Atoms for Scalable Text Analysis in Data Analytics\\
    in the Era of LLM-Based Agents}

  \begin{icmlauthorlist}
    \icmlauthor{Jinxiang Xie}{nju}
    \icmlauthor{Zihao Li}{thu}
    \icmlauthor{Rui Ding}{msr}
    \icmlauthor{Bartuer Zhou}{msr}
    \icmlauthor{Shi Han}{msr}
    \icmlauthor{Dongmei Zhang}{msr}
  \end{icmlauthorlist}

  \begin{center}
    \small
    \textsuperscript{1}Nanjing University\quad
    \textsuperscript{2}Tsinghua University\quad
    \textsuperscript{3}Microsoft Research
  \end{center}

  \icmlaffiliation{nju}{Nanjing University}
  \icmlaffiliation{thu}{Tsinghua University}
  \icmlaffiliation{msr}{Microsoft Research}

  \icmlcorrespondingauthor{Jinxiang Xie}{\nolinkurl{xiejinxiang@smail.nju.edu.cn}}
  \icmlcorrespondingauthor{Rui Ding}{\nolinkurl{juding@microsoft.com}}

  \icmlkeywords{LLM agents, semantic operators, text analytics, data analytics}

  \vskip 0.3in
]

\printAffiliationsAndNotice{}

\begin{abstract}
Much of the information needed for data analysis is buried in text
columns: an analyst may infer customer frustration from complaint text and
use it to explain resolution time. We call workloads in which processing
text columns is central to subsequent data analysis \sys{}
(\emph{Text Analysis in Data Analytics}). In an agentic setting, the user's
query dynamically specifies this processing, and answering it interleaves
LLM-based semantic steps with exact SQL or Python operations. This mixed
execution raises two design questions: which work should be neural and which
symbolic, and which semantic outputs later steps can reuse.
We address them for agent-planned \sys{} workflows with two complementary,
query-driven semantic atoms. Each atom takes a sequence of text items and a
natural-language query specifying the semantic task: \textbf{Tagging}
returns one string per input item ($N\!\to\!N$), while
\textbf{Categorization} returns $M$ category objects over $N$ input items
($N\!\to\!M$). Tagging results are materialized as a row-aligned column and
Categorization results as a category table; SQL or Python then performs exact
computation over this exposed state. The atoms add no expressive power over
semantic-operator systems that offer a semantic map; Tagging is functionally
such a map. Their purpose is to make reusable semantic state the uniform,
default output of every semantic call. We realize this interface in
TADA-Agent, whose atoms run inside generated Python programs and internally
handle inputs beyond a reliable context window. We evaluate TADA-Agent on
TABench, a benchmark of 40 complete analysis requests over spreadsheet tables
spanning workflow composition, input--output structure, and
literature-grounded analytical scenarios. Against a semantic-operator baseline
and relative to a held-out TADA reference run, TADA-Agent delivers comparable
row-level semantic quality, has higher complete-artifact preservation
estimates, and uses over an order of magnitude fewer semantic-model requests
under the evaluated native execution paths. Large-scale fixed-taxonomy runs
and public legal and biomedical benchmarks show that the same Tagging API can
be used beyond one context window and across domains, with task-specific
queries and downstream normalization.
\end{abstract}

\section{Introduction}

Text has long been part of analytical data, yet it has seldom been handled as
a core analytical substrate in its own right. The relational model gives
analysts a clean algebra over structured columns~\citep{codd1970relational},
while classical text analysis provides a separate toolbox for turning
documents into linguistic features, labels, or summaries
~\citep{manning1999foundations}. Modern analytical tables, however,
increasingly combine text and structured columns. For example, a
customer-support table may contain a text column of billing complaints
alongside structured columns such as resolution time and region. We refer to
such tables as \emph{text-bearing tables}. An analyst may infer each
complaint's frustration and main theme from the text, then use these derived
variables in a regression to explain variation in resolution time.
We use \sys{} (\emph{Text Analysis in Data Analytics}) for such workloads, in
which semantic processing over text columns produces intermediate artifacts
for subsequent data analysis. Throughout the paper, \sys{} names this
workload, TADA-Agent names our system, and Tagging and Categorization name the
interface that the system exposes to an agent.

The execution of \sys{} requests is changing in the agentic era. Semantic-query
interfaces now support both human-authored programs and agentic execution. In
an LLM-based analytical agent, the user expresses analytical intent through a
natural-language query, and the agent may plan a multi-step
execution to answer it~\citep{yao2023react,tang2025dataanalystsurvey}. LLM
calls perform the \emph{neural computation} required for semantic processing
over text, while SQL, dataframe, and statistical runtimes provide
\emph{symbolic computation} for exact filtering, sorting, aggregation, and
model fitting. Within such an execution, a semantic result may become an
intermediate variable reused by later semantic or symbolic steps. In short,
\sys{} in the agentic setting has two salient characteristics: semantic
needs are specified dynamically by the query rather than fixed by a
predefined task, and answering the query may require neural and symbolic
steps to be interleaved. Consider the query, ``Sort the billing complaints by
customer frustration, identify the main theme of each complaint, and regress
resolution time on frustration and complaint theme.'' It dynamically
specifies frustration and theme as semantic variables, which must be derived
through neural processing and then used by symbolic sorting and regression,
raising two challenges.

The first challenge is interface design. The system must decide which work is
done by an LLM and which by exact code, and which LLM outputs are kept as
\emph{semantic state}: values, such as a frustration score per row, that later
steps can read without calling the model again. TADA requires both neural and symbolic computation: LLM calls are needed
to perform query-specified semantic processing over text, whereas operations
such as applying a filter, sorting values, computing aggregates, and fitting
models have exact symbolic implementations. A poorly placed boundary can reduce
semantic coverage or execution reliability. Symbolic surrogates are often
inadequate for context-dependent meanings, while delegating operations with
established deterministic implementations directly to generation introduces
avoidable variability. In the running
example, sorting by frustration should use neural computation to infer a
row-level frustration value and symbolic code to perform the actual sorting.

Existing systems often expose \emph{task-level semantic operators}, each of
which packages semantic processing with a particular analytical action, such
as filtering, sorting, or aggregation
~\citep{snowflake2026aisql,databricks2026aifunctions,doris2026aifilter,patel2025lotus,liu2024suql,dai2024uqe,lee2025saber,qi2026sema}.
A composite contract determines which state the orchestrator can reuse. The
orchestrator can use only the state exposed by an operator's output
contract. In the billing-complaint example, a semantic sort returns
ordered complaints but not the row-level frustration variable required by the
later regression. A plan that relies on this sort must invoke an additional
semantic computation to derive that variable, repeating the semantic
processing for the same notion of frustration and potentially producing values
inconsistent with the earlier ordering, as illustrated in
Figure~\ref{fig:interface-comparison}(a). The repetition stems from this
plan choice. Most semantic-operator systems also provide a
semantic map, and a plan that maps frustration first and then sorts and
regresses symbolically avoids the repetition. Whether the agent finds that
plan depends on which operators it is offered and which outputs those
operators return by default. The system must therefore place
two boundaries well. The \emph{execution boundary} decides which work is
neural and which is symbolic; the \emph{interface boundary} decides what
semantic state each call exposes to subsequent analytical steps.

The second challenge is scale: a large table may not fit reliably in one call.
Consider identifying the main topics in
thousands of complaints. A natural divide-and-conquer strategy follows a
MapReduce structure: partition the rows, process each chunk, and consolidate
the local results. Yet each stage raises questions specific to semantic
computation. During partitioning, large chunks approach the model's input and
output limits, whereas small chunks provide less evidence and may produce
unstable local topics. During local processing, the system must preserve information that is
infrequent within one chunk but potentially important globally. Splitting
and reassembling rows is not enough: the system must select suitable
chunks, preserve globally important semantics through local LLM calls, and
consolidate the local results reliably.

To address the interface-design challenge, TADA places the neural--symbolic execution
boundary at the semantic output and exposes that output through two
complementary, query-driven atoms. \textbf{Tagging} maps $N$ input text
items, typically drawn from a text column, to $N$ output strings
($N\!\to\!N$), while \textbf{Categorization} maps the same $N$ text items to
a list of $M$ category objects ($N\!\to\!M$). Both atoms also take a query
$q$ that specifies the semantic processing to perform. This atom-level query
need not be identical to the user's
analytical query: the agent may derive it as one step in the workflow. The
semantic task is therefore query-defined, while the two output contracts
remain fixed. LLM
calls perform the semantic processing needed to produce the atom outputs;
filtering, sorting, and model fitting remain in the symbolic
runtime. Tagging values are materialized as a row-aligned column keyed by
row identifiers (\texttt{rid}), whereas Categorization objects are
materialized as a separate category table keyed by fresh category
identifiers (\texttt{cid}).
The boundary therefore both confines neural computation to semantic processing
and exposes its result as state that subsequent analytical steps can reuse.
The TADA plan in Figure~\ref{fig:interface-comparison}(b) Tags frustration
once: symbolic code sorts by the materialized column, and the regression
reuses that same column. Categorization discovers a shared theme vocabulary,
and a subsequent Tagging call maps each complaint to the most relevant
discovered theme. This Categorization--then--Tagging composition provides the
discovery-and-assignment functionality of a task-level semantic grouping
operation while preserving both the category vocabulary and its row-level
assignments: LLM calls produce semantic state, exact operations consume it,
and that state remains available across subsequent steps. The two atoms do
not compute anything that a semantic-operator system with a semantic map
cannot also compute: Tagging is functionally such a map. What TADA changes is
the default. Every semantic call returns reusable state, so the map-first
decomposition is the plan the interface offers by default, and the agent does
not have to discover it among composite alternatives.

\begin{figure}[t]
  \centering
  \begin{minipage}{\columnwidth}
    \small\textbf{(a) Composite semantic operators}\\[2pt]
    \begin{lrbox}{\codebox}\begin{minipage}{\dimexpr\linewidth-2\fboxsep-2\fboxrule\relax}
\scriptsize
\begin{alltt}
ordered = semantic_sort(text, "frustration")
theme = semantic_theme(text)
\textcolor{red!70!black}{# frustration is processed again}
frustration = semantic_map(text, "frustration")
regress(resolution_time \(\sim\) frustration + theme)
\end{alltt}
    \end{minipage}\end{lrbox}\fbox{\usebox{\codebox}}\\[6pt]
    \textbf{(b) TADA atoms and symbolic operations}\\[2pt]
    \begin{lrbox}{\codebox}\begin{minipage}{\dimexpr\linewidth-2\fboxsep-2\fboxrule\relax}
\scriptsize
\begin{alltt}
frustration = ta.tag(text, "score frustration 1-10")
ordered = sort(table, by=frustration)
themes = ta.categorize(text, "main complaint themes")
theme = ta.tag(text, "map to a label in \{themes\}")
regress(resolution_time \(\sim\) frustration + theme)
\end{alltt}
    \end{minipage}\end{lrbox}\fbox{\usebox{\codebox}}
  \end{minipage}
  \caption{Two plans for the same user query: ``Sort the billing complaints
  by customer frustration, identify the main theme of each complaint, and
  regress resolution time on frustration and complaint theme.''
  \textbf{(a)}~A plan built from composite semantic operators receives only
  each step's final result: the sort exposes an ordering but no frustration
  values, so the regression derives frustration again. Moving the map call
  before the sort would avoid this; the figure shows the plan that a composite
  sort invites, not the only plan an operator system permits. \textbf{(b)}~TADA Tags
  frustration once as a reusable column that symbolic sorting and regression
  both consume; Categorization then Tagging yields a shared theme vocabulary
  together with its row-level assignments.}
  \Description{For a query that sorts billing complaints by frustration,
  identifies each complaint's main theme, and regresses resolution time on
  frustration and theme, side-by-side pseudocode compares composite semantic
  operators with TADA atoms. The composite plan derives frustration again
  after sorting, whereas TADA reuses one frustration column for sorting and
  regression.}
  \label{fig:interface-comparison}
\end{figure}

To address the second challenge, the two atom APIs handle large-scale
execution internally, so the agent uses the same interface at every input
scale. Large-scale Categorization follows three stages. Planning fixes a
chunk size $n$ under model capacity and uses a pilot to select the local
retention budget $r\geq k$ for the requested category count $k$. Local
processing randomly partitions the rows and retains up
to $r$ candidates per chunk, preserving more information than local top-$k$
alone. A semantic reduction then consolidates synonymous and overlapping
candidates into the final category table. We formulate local retention as a
top-$k$ preservation problem: a sufficient condition identifies how chunk
size, retention, and support margins affect recovery, and a theory-guided
bootstrap selects the smallest $r$ meeting a target reliability.

Large-scale Tagging follows two paths. Independent Tagging partitions rows,
applies the same query to each chunk, and reassembles outputs by
\texttt{rid}. Joint Tagging is needed when rows must share an output domain:
large-scale Categorization first establishes one category list, after which
row assignment reduces to the independent path. Partitioning, validation,
parallel calls, retries, and consolidation remain internal to the two atoms.

We realize the interface in TADA-Agent, an analytical agent that generates
and executes Python workflows over text-bearing tables. \texttt{ta.tag} and
\texttt{ta.categorize} are Python APIs invoked through Programmatic Tool
Calling~\citep{anthropic2025advancedtooluse}, so atom calls remain inside
the same runtime as dataframe operations and control flow. In contrast,
exposing the atoms only as peer tools through Traditional Tool Calling would
unfold a data-dependent loop into repeated orchestrator--tool interactions
and move intermediate tables and category lists across that boundary. A
strong reasoning model plans the analysis while efficient lightweight models
execute focused atom calls. A text-analysis skill supplies API contracts,
composition patterns, and canonical examples, but the orchestrator still
chooses the workflow and symbolic methods.

We evaluate the interface on TABench, 40 complete analysis requests over
spreadsheet tables, organized by workflow composition, input--output structure, and a
literature-grounded scenario taxonomy. TADA-Agent and semantic operators are
close on classification and extraction; semantic operators better reproduce
the reference output's absolute ranking scale, whereas TADA-Agent's ordering
agrees more with the reference and it performs best on constrained generation.
TADA-Agent also has higher complete-artifact preservation estimates than
semantic operators in all five evaluated classes, although a coding agent
without the atoms matches these estimates in three of the five.
These comparisons are made against a held-out TADA run as the common
reference, and the class-level intervals overlap, so they describe agreement
with that reference rather than absolute quality; in particular, ordering
agreement partly reflects TADA-Agent's run-to-run consistency.
Under the evaluated native execution paths, it uses at least 41.8 times fewer
dedicated semantic-model requests than the semantic-operator implementation;
this measures implementation-level request fan-out rather than an intrinsic
property of every operator interface. The large-scale evaluation covers both
atoms at different levels: Tagging is
evaluated end to end beyond one context window, while a controlled study of
Categorization isolates whether locally retained candidates recover the global
top-$k$ under fixed assignments. In this study, $r=k$ suffices when the cutoff
is separated, whereas near-tied distributions require a larger pilot-selected
budget. Public legal and biomedical benchmarks separately test whether the
unchanged Tagging API transfers beyond TABench with benchmark-specific
postprocessing.

Our contributions are:

\noindent\textbf{1. An agent-era formulation and semantic-atom interface.}
We frame agent-driven \sys{} around two design decisions: which work an LLM
performs and which exact code performs, and which LLM outputs remain available
to later steps. We identify two recurring shapes of such reusable outputs:
row-aligned values and corpus-level vocabularies. We expose these shapes through
complementary $N\!\to\!N$ Tagging and $N\!\to\!M$ Categorization atoms. They
add no expressive power over semantic operators; the contribution is making
reusable state the default output of every semantic call. They are an initial interface,
not a universal algebra.

\noindent\textbf{2. TADA-Agent and scalable atom execution.}
TADA-Agent invokes the atoms from generated Python and preserves their public
contracts across one-call and multi-call execution.

\noindent\textbf{3. TABench and empirical evaluation.}
TABench provides 40 complete \sys{} requests spanning workflow, structure,
and scenario. The evaluation separates row-level quality, artifact
reliability, and semantic-model request cost, with additional evidence on
large-scale execution and public extraction benchmarks.

\section{Related Work}

\subsection{Semantic Operators and AI Functions}

Semantic query processing is the closest line of work to TADA. LOTUS provides
both LLM and tool-using agentic semantic operators; its dataframe APIs cover
mapping, filtering, extraction, aggregation, ranking, and joins, with optimized
implementations and accuracy guarantees~\citep{patel2025lotus}. SUQL extends SQL with free-text primitives
for question answering and summarization over unstructured passages
~\citep{liu2024suql}; UQE compiles a SQL-like language for unstructured data
using LLM reasoning, sampling, and classical query optimization
~\citep{dai2024uqe}. Sema couples a high-performance DuckDB engine with a
semantic query algebra~\citep{qi2026sema}; SABER proposes an extended
relational algebra as a unifying, SQL-compatible foundation
~\citep{lee2025saber}. Industrial platforms expose related AI functions in SQL
~\citep{snowflake2026aisql,databricks2026aifunctions,doris2026aifilter}.
Across these interfaces, natural language specifies semantic processing, and
most of them already separate model-backed calls from relational execution.
TADA does not add expressive power to this design space: its Tagging atom is
functionally a semantic map. It differs in which outputs the agent receives by
default. Every semantic call returns either a row-aligned column or a category
table, exact operations remain in the symbolic runtime, and the orchestrator
composes both through generated code.

The implementations reveal an execution boundary consistent with TADA's
design. LOTUS semantic filtering produces Boolean decisions and applies the
resulting mask to a dataframe, while semantic top-$k$ places a pairwise model
comparator inside deterministic selection and sorting algorithms
~\citep{patel2025lotus}. Snowflake and Apache Doris similarly expose AI
filters as Boolean functions used by an ordinary SQL \texttt{WHERE} clause
~\citep{snowflake2026aifilter,doris2026aifilter}. Thus a logical API may be a
composite semantic operator even when its implementation already embeds
semantic processing in a symbolic algorithmic shell. TADA makes this boundary
explicit and uniform across its two atoms: LLM calls produce semantic values,
and the surrounding runtime performs selection, sorting, and aggregation.

What each operator returns also differs by operator: an
operator may return a filtered or top-$k$ dataframe, a joined table, or a
textual aggregate. LOTUS, for example, returns selected rows from
\texttt{sem\_topk} and text from \texttt{sem\_agg} unless the caller supplies
a structured format~\citep{patel2025lotus}. Existing systems also expose
useful escape hatches, including LOTUS filter decisions and raw outputs,
Databricks extraction schemas, and DocETL intermediate reductions
~\citep{patel2025lotus,databricks2026aifunctions,shankar2025docetl}. These
mechanisms are optional and operator-specific. SABER independently identifies
the lack of uniform semantics and composability as a systems problem
~\citep{lee2025saber}. TADA's difference from these systems is a design
choice, not a capability gap: it makes two uniformly shaped outputs the primary
interface: row-level values remain available to subsequent symbolic
operations, and category objects give later Tagging calls a shared, reusable
vocabulary.

\subsection{Analytical Agents and Semantic Pipeline Systems}

Tool-using LLM agents provide the execution setting that motivates the TADA
interface. ReAct interleaves reasoning and actions
~\citep{yao2023react}, while Toolformer studies self-supervised learning of
which APIs to call, when, and with what arguments~\citep{schick2023toolformer}.
Recent surveys describe agents that
orchestrate SQL, code, visualization, and LLM calls for data analysis
~\citep{tang2025dataanalystsurvey}. Programmatic Tool Calling further allows
generated code to invoke model-backed tools inside a code-execution
environment, avoiding a model round trip for every iteration of a loop
~\citep{anthropic2025advancedtooluse}. TADA-Agent uses this execution pattern
to place Tagging and Categorization inside the same Python workflow as
dataframe and statistical operations.

Several semantic data systems target whole pipelines rather than single
operators. DocETL represents
complex document-processing pipelines and uses agentic rewrites, plan
evaluation, and optimization to improve their execution
~\citep{shankar2025docetl}. Palimpzest searches over logical and physical
plans to trade off quality, cost, and runtime
~\citep{liu2025palimpzest}. ZenDB extracts semantic document structure for
efficient downstream queries~\citep{lin2024zendb}, while EVAPORATE constructs
structured views over heterogeneous data lakes
~\citep{arora2023evaporate}. TAG argues for database-backed generation beyond
Text2SQL and retrieval-only interfaces~\citep{biswal2024tag}. These systems
address the same overall problem of representing, optimizing, and answering
complex data requests, usually from a declarative program or plan. TADA
approaches it from the side of an agent that writes the plan step by step,
and studies one part of it: what a semantic call should return when its result
must remain available to later filtering, ranking, grouping, comparison, and
statistical modeling.

\paragraph{Benchmarks.} SemBench evaluates semantic query processing engines
such as LOTUS, Palimpzest, ThalamusDB, and BigQuery on SQL queries with
semantic operators across scenarios and data modalities, measuring cost,
latency, and quality against labels derived from human annotations
~\citep{lao2026sembench}. TABench is complementary rather than a replacement.
It is text-only and smaller, and each case is a complete analysis request
posed to an agent in natural language, so the agent must choose the
decomposition, combine semantic and exact steps, and deliver output files
under an explicit contract. Many of these requests have no unique answer,
which is why our evaluation uses a reference run and an LLM judge; SemBench's
human-label ground truth avoids this limitation for the operator-level
queries it covers.

\subsection{Stable and Large-Scale Semantic Processing}

TADA-Agent also builds on work that improves the reliability and scale of
LLM-based data processing. CAST studied corpus-level Summarization and row-level Tagging, including their
run-to-run stability under
Algorithmic Prompting and Thinking-before-Speaking~\citep{xie2026cast}.
Building on these prompting principles, TADA defines Categorization and
Tagging as query-driven $N\!\to\!M$ and $N\!\to\!N$ contracts for
agent-generated analytical workflows.

Large inputs require more than independent LLM calls. DocETL uses
hierarchical reduction and optimization for long-document workloads
~\citep{shankar2025docetl}; UQE combines sampling with compiler-style
execution~\citep{dai2024uqe}; and semantic query systems optimize batching,
operator order, and model invocation
~\citep{patel2025lotus,qi2026sema}. TADA focuses specifically on preserving
the two atom outputs across multiple calls: independent Tagging is reassembled
by row identity, Categorization semantically consolidates local proposals,
and joint Tagging follows discover--then--assign.

The top-$k$ analysis draws on classical distributed ranking and concentration
results. Fagin, Lotem, and Naor study instance-optimal aggregation algorithms
for top-$k$ middleware queries~\citep{fagin2003optimal}; Theobald, Weikum, and
Schenkel analyze top-$k$ evaluation with probabilistic guarantees
~\citep{theobald2004topk}; and Hoeffding-style bounds control ranking
inversions under randomized sampling~\citep{hoeffding1963probability}.
TADA applies these ideas to local candidate retention in a semantic
Categorization pipeline. The controlled analysis fixes canonical assignments
to isolate retention, while the end-to-end scale experiment evaluates
fixed-taxonomy Tagging.

\section{Two Semantic Atoms and Their Composition}
\label{sec:abstraction}

This section defines the two semantic atoms, Tagging and Categorization, and
presents a systematic taxonomy of their composition with symbolic operations.
We first recover familiar semantic operators and then examine broader
analytical workflows enabled by the same atom outputs.

\subsection{Tagging and Categorization}
\label{sec:two-atoms}

Both atoms accept a sequence of text items and an atom-level query. Let
$X=(x_1,\ldots,x_N)$ denote the $N$ input texts, and let $q$ specify the
semantic processing to perform. The texts
typically come from one table column. When the requested processing depends on
several columns, the agent first serializes their row-wise values into an
intermediate text item $x_i$. The
atom-level query $q$ may likewise be derived from the user's analytical query
as part of the agent's plan.

\paragraph{Tagging.}
Tagging returns one string for each input text:
\begin{equation}
    \textsc{Tag}(X,q)
    \rightarrow
    S=(s_1,\ldots,s_N).
    \label{eq:tagging}
\end{equation}
Each $s_i$ is the result for the corresponding input text $x_i$. Depending on
$q$, the string may contain a sentiment label, an extracted postcode, a
keyword, a summary, or a numeric score written as text. The output list
preserves the length and order of $X$. When the result is materialized,
TADA-Agent writes the list as one new column, associating each $s_i$ with
the \texttt{rid} of its input row.

The atom query specifies the semantic processing. For example, it may ask
Tagging to classify sentiment using a given label set, extract a postcode, or
assign a score from 1 to 10. A request for several output columns calls
Tagging once per column. Extracting province, city, and postcode therefore
uses three calls with three queries. Each call keeps the same $N\!\to\!N$
form in
Equation~\ref{eq:tagging}.

\paragraph{Categorization.}
Categorization returns a list of category objects for the input texts:
\begin{equation}
    \textsc{Categorize}(X,q)
    \rightarrow
    \mathcal C=(c_1,\ldots,c_M).
    \label{eq:categorize}
\end{equation}
Each $c_j$ contains a required name and an optional description. The atom
query specifies how the texts should be categorized (for example, by topic
or issue type) and any constraints on the categories, including their
number. The returned count $M$ is the number of objects produced for that
request. When materialized, TADA-Agent writes the objects as a separate
category table and assigns a fresh \texttt{cid} to each row.

Categorization produces the category objects but does not assign them to
input rows. A workflow that needs both operations first calls Categorization,
inserts the returned names and descriptions into a new atom query, and then
calls Tagging. The new query may, for example, ask Tagging to ``map each
review to the most relevant label in the following list.'' Category discovery
and row assignment are thus two explicit atom calls linked by query text.
\subsection{From Semantic Operators to Analytical Workflows}
\label{sec:composition}

Tagging and Categorization expose deliberately simple outputs: one string per
row and one list of categories, respectively. Their usefulness comes from the
analytical workflow that composes them. TADA-Agent can construct a
sequence of atom-level queries from the user's analytical query, invoke either
atom, materialize its output, and continue with another atom call or a
symbolic operation. Repeated calls create several columns; row-wise
serialization brings several input columns into one Tagging call; loops apply
the same atom to groups or time periods; and materialized labels, scores, and
categories become inputs to OLAP operations, control flow, and model fitting.
The frustration and theme columns in
Figure~\ref{fig:interface-comparison}(b) illustrate this reusable state in one
multi-step workflow.

Many familiar semantic operators are derived workflows over these outputs.
A semantic map is a Tagging call whose result is written as a column. A
semantic filter Tags a Boolean decision and applies an ordinary Filter. A
pointwise semantic sort Tags the requested row-level value and invokes an
ordinary Sort, while a top-$k$ endpoint adds a symbolic cutoff. Fixed-label
semantic grouping Tags a group name and applies GroupBy. Open grouping first
Categorizes the column to discover a shared vocabulary, then Tags every row
under that vocabulary before GroupBy. Whole-column summarization is a
Categorization query requesting one summary. These decompositions mirror the
physical structure of well-engineered semantic operators, and the same
workflows can be written with an operator system that exposes a semantic map.
The difference is that the neural output is the primary analytical artifact by
default, whereas in a composite operator it is an internal or optional value
behind one action.

Table~\ref{tab:composition-taxonomy} moves from the single example in
Figure~\ref{fig:interface-comparison} to 16 representative requests. Its
scenario axis comes from the
literature-grounded taxonomy used to construct TABench. The taxonomy
synthesizes established work on classification and extraction
~\citep{hendrycks2021cuad,doosterlinck2023biodex,arora2023evaporate,xie2026cast},
semantic filtering, grouping, aggregation, and ranking
~\citep{patel2025lotus,dai2024uqe,liu2024suql,qi2026sema,lee2025saber,lao2026sembench},
and multi-step document and analytical workflows
~\citep{shankar2025docetl,lin2024zendb,liu2025palimpzest,tang2025dataanalystsurvey}.
A request may carry several scenario labels because one analytical workflow
can perform, for example, classification, filtering, and summarization in
sequence.

\begin{table*}[t]
\centering
\caption{Representative TADA requests viewed by scenario, input--output
structure, workflow composition, and correspondence to existing semantic
operators. Scenario names follow the fine-grained intent taxonomy; the
appendix groups these leaves into 12 broader reporting classes.}
\label{tab:composition-taxonomy}
\begingroup
\setlength{\tabcolsep}{3pt}
\renewcommand{\arraystretch}{1.06}
\scriptsize
\begin{tabular}{
>{\raggedright\arraybackslash}p{0.205\textwidth}
>{\raggedright\arraybackslash}p{0.13\textwidth}
>{\raggedright\arraybackslash}p{0.15\textwidth}
>{\raggedright\arraybackslash}p{0.33\textwidth}
>{\raggedright\arraybackslash}p{0.11\textwidth}}
\toprule
\textbf{Representative request}
& \textbf{Scenario}
& \textbf{Input--output}
& \textbf{TADA workflow}
& \textbf{Existing operator} \\
\midrule
\multicolumn{5}{l}{\textit{Workflows corresponding to familiar semantic-operator families}} \\
\addlinespace[2pt]
Label each review as positive, negative, or neutral
& Classification
& $N{\times}1\to N{\times}1$
& Tag with the three labels written in the query
& \texttt{sem\_map} \\
\addlinespace
Discover issue categories and group the reviews by them
& Grouping; classification
& $N{\times}1\to(M,N{\times}1)$
& Categorize $\to$ build a query from the category names $\to$ Tag $\to$
GroupBy
& Semantic grouping \\
\addlinespace
Keep comments that imply a risk of customer churn
& Semantic filtering
& $N{\times}1\to N'{\times}1$
& Tag a \texttt{Yes}/\texttt{No} decision $\to$ Filter
& \texttt{sem\_filter} \\
\addlinespace
Extract time, location, scope, and cause from each incident
& Information extraction
& $N{\times}1\to N{\times}w$
& Invoke Tagging $w$ times, once for each output column
& \texttt{sem\_map}/ \allowbreak\texttt{sem\_extract} \\
\addlinespace
Produce one overall summary of the survey responses
& Text summarization
& $N{\times}1\to1$
& Categorize with a query requesting one overall summary
& \texttt{sem\_agg} \\
\addlinespace
Return the ten support tickets with highest inferred urgency
& Semantic ranking
& $N{\times}1\to k$ rows
& Tag a score $\to$ convert strings to numbers $\to$ Sort $\to$ take $k$
& \texttt{sem\_topk} \\
\addlinespace
Map country-name variants to a canonical country name
& Data quality repair
& $N{\times}1\to N{\times}1$
& Tag the canonical name
& \texttt{sem\_map} \\
\addlinespace
Infer priority from description, category, and customer tier
& Classification; column generation
& $N{\times}y\to N{\times}1$
& Serialize the $y$ input values row by row $\to$ Tag a priority
& \texttt{sem\_map} \\
\midrule
\multicolumn{5}{l}{\textit{Workflows composed beyond one semantic operator}} \\
\addlinespace[2pt]
Discover the main feedback topics for each organization
& Text summarization
& $\{N_g{\times}1\to M_g\}_{g\in G}$
& GroupBy organization $\to$ ForEach group: Categorize
& No single operator \\
\addlinespace
Discover broad topics and then subtopics within each topic
& Grouping; Summarization
& $N{\times}1\to(M_1,\{M_{2,j}\})$
& Categorize $\to$ build a query $\to$ Tag $\to$ GroupBy $\to$
ForEach group: Categorize
& No single operator \\
\addlinespace
Discover the main topics among negative reviews
& Classification; Filtering; Summarization
& $N{\times}1\to N'{\times}1\to M$
& Tag sentiment $\to$ Filter negative rows $\to$ Categorize
& No single operator \\
\addlinespace
Compute the average rating of semantically negative reviews
& Classification; Filtering; text--numeric analysis
& $N{\times}(\text{text},\text{number})\to1$
& Tag sentiment $\to$ Filter negative rows $\to$ Average rating
& No single operator \\
\addlinespace
Explain products with unusually high return rates from their comments
& Summarization; text--numeric analysis
& $N{\times}(\text{text},\text{number})\to N'{\times}\text{text}\to M$
& Detect high return rates $\to$ Filter products $\to$ Categorize comments
& No single operator \\
\addlinespace
Rate reviews from 1 to 5 with only the top 10\% receiving 5
& Scoring; Ranking
& $N{\times}1\to N{\times}1$
& Tag raw scores $\to$ convert to numbers $\to$ Sort $\to$ compute cutoffs
$\to$ write final ratings
& No single operator \\
\addlinespace
Find topics appearing this month but absent from earlier months
& Summarization; pattern detection
& $\{N_t{\times}1\to M_t\}_{t\in T}$
& GroupBy time $\to$ ForEach period: Categorize $\to$ compare category lists
& No single operator \\
\addlinespace
Use discovered review aspects as features in a regression
& Grouping; column generation
& $N{\times}1\to M\to N{\times}M\to$ model
& Categorize aspects $\to$ Tag once per aspect $\to$ Regression
& No single operator \\
\bottomrule
\end{tabular}
\endgroup
\end{table*}

Table~\ref{tab:composition-taxonomy} spans all nine supported taxonomy
families, from classification and semantic grouping to pattern detection,
data repair, and semantic column generation. Several requests combine these
families: finding topics among negative reviews composes classification,
filtering, and summarization; emerging-topic analysis combines summarization
with temporal pattern detection.

The table also covers output structures that are not obvious from either atom
alone. Inputs range from one text column to several columns serialized by row.
Outputs include one or many derived columns, filtered row subsets, category
lists with row assignments, group-specific lists, hierarchical categories,
scalar aggregates, ordered rows, and fitted models. Hierarchical topic
analysis recursively invokes Categorization over groups created from an
earlier result. Semantic feature construction discovers aspects, converts
each aspect into a row-aligned predictor, and passes those predictors to a
statistical model. Thus the fixed $N\!\to\!N$ and $N\!\to\!M$ atom contracts
do not fix the shape of the complete analytical result.

The atom query determines how a returned string is interpreted by the next
operation: as a Boolean decision for Filter, a category name for GroupBy, a
number for Sort or Average, or text retained as a new column.

The first eight rows of Table~\ref{tab:composition-taxonomy} recover familiar
semantic-map, filter, grouping, aggregation, and ranking families. Each uses
one or two atom calls, optionally followed by symbolic code. The remaining
rows cross these operator boundaries. They include loops over groups,
hierarchical discovery, semantic filtering followed by summarization,
text-conditioned numeric aggregation, temporal comparison, global score
constraints, and regression over semantically constructed features. The
interface does not need a separate model-backed operator for each analytical
verb; the agent builds these workflows from two exposed output shapes.

The constrained-rating request illustrates the split. Tagging scores every
review from 1 to 5. The symbolic runtime converts the strings to numbers,
sorts them, computes percentile cutoffs, and rewrites the scores so that only
the top 10\% receive 5. The LLM supplies row-level semantics; symbolic code
enforces the global constraint exactly. Control flow can likewise apply
Categorization once per organization or time period, or recursively within
every discovered topic. These patterns keep semantic outputs available to the
rest of the workflow.

\section{TADA-Agent Design and Implementation}
\label{sec:agent-design}

Section~\ref{sec:abstraction} defines the semantic state that an analytical
workflow can reuse. TADA-Agent realizes this interface by making Tagging and
Categorization available inside its Python code-execution environment. Three
design choices are central: Programmatic Tool Calling keeps atom calls inside
generated control flow, separate model roles assign orchestration and atom
execution to models suited to each kind of work, and a text-analysis skill
teaches the orchestrator how to construct valid workflows.

\subsection{Programmatic Tool Calling for Text Analysis}
\label{sec:programmatic-tool-calling}

Under \emph{Traditional Tool Calling}, an orchestrator invokes a top-level tool, receives its result, and then chooses the next action. Tagging and Categorization could be registered as tools beside a Python tool. This design is natural for isolated semantic operations, but it places the orchestrator between every semantic call and the code around it.

TADA workflows frequently invoke semantic processing inside data-dependent
OLAP operations and control flow. Consider ``for each organization, summarize
the topics in its reviews.'' With Traditional Tool Calling, the orchestrator
must enumerate the organizations, return from Python execution, invoke a
semantic tool for every organization, retain each result, and assemble the
mapping. The compact loop is unfolded into a sequence of model decisions and
tool calls. Group-specific tables and partial results repeatedly cross the
model--tool boundary, increasing round trips and creating opportunities to
omit a group or lose the correspondence between a group and its result.

Anthropic uses \emph{Programmatic Tool Calling} for the alternative in which
generated code invokes tools inside a code-execution environment
~\citep{anthropic2025advancedtooluse}. TADA-Agent follows this pattern:
\texttt{ta.tag} and \texttt{ta.categorize} are Python APIs callable from the
generated workflow. The group-wise request becomes:

\begin{figure}[t]
  \centering
  \begin{minipage}{\columnwidth}
    \small\textbf{(a) Traditional tool calling}\\[2pt]
    \begin{lrbox}{\codebox}\begin{minipage}{\dimexpr\linewidth-2\fboxsep-2\fboxrule\relax}
\scriptsize
\begin{alltt}
orgs = python(groupby)  \textcolor{red!70!black}{# orgs enter context}
for org in orgs:
    rows = python(slice)  \textcolor{red!70!black}{# rows enter context}
    topics = categorize(rows)
    \textcolor{red!70!black}{# topics enter context}
assemble(topics)
\end{alltt}
    \end{minipage}\end{lrbox}\fbox{\usebox{\codebox}}\\[6pt]
    \textbf{(b) Programmatic tool calling}\\[2pt]
    \begin{lrbox}{\codebox}\begin{minipage}{\dimexpr\linewidth-2\fboxsep-2\fboxrule\relax}
\scriptsize
\begin{alltt}
topics = \{\}
for org, rows in table.groupby("org"):
    topics[org] = ta.categorize(rows)
    \textcolor{teal!70!black}{# state stays in runtime}
write_json(topics) \textcolor{teal!70!black}{# final result only}
\end{alltt}
    \end{minipage}\end{lrbox}\fbox{\usebox{\codebox}}
  \end{minipage}
\caption{Group-wise Categorization. \textbf{(a)}~Traditional tool calling
unfolds the loop into orchestrator steps, repeatedly placing groups, table
slices, and partial results in model context. \textbf{(b)}~Programmatic Tool
Calling keeps the loop and intermediate state inside the executed workflow;
only the final mapping returns.}
\Description{Side-by-side pseudocode contrasts traditional tool calling, in
which organizations, row slices, and topic results cross the model boundary,
with Programmatic Tool Calling, in which the loop and intermediate state
remain inside code execution.}
\label{fig:groupwise-pseudocode}
\end{figure}

The organizations are discovered at runtime, while dataframe slices and
category lists remain inside code execution. The same interface supports
deeper nesting: hierarchical topic analysis Categorizes the full column, Tags
rows with the discovered names, and Categorizes each resulting group.
Multi-output extraction invokes Tagging several times; constrained scoring
Tags a semantic value and lets numpy or pandas enforce global cutoffs. The
atoms therefore remain ordinary calls inside the workflow rather than an
API vocabulary that grows with each composition.

Programmatic Tool Calling assumes a safe code-execution environment and
couples execution to the libraries available there. TADA-Agent makes this
trade-off because its target workflows already require dataframe,
statistical, and machine-learning libraries. Keeping semantic calls in the
same runtime preserves loop variables, table state, exceptions, and result
structure across the whole analysis workflow.

\subsection{Separate Model Roles}
\label{sec:model-roles}

The orchestrator and atom calls solve different problems. The orchestrator
analyzes the request, reasons over the schema, chooses a decomposition, writes
the workflow, and responds to execution failures. This role benefits from a
strong reasoning model because its decisions span the complete analysis.

An atom call has a narrower contract. The library combines a fixed system
prompt with one query and a text column, invokes the selected model, validates
the structured response, and returns one of two prescribed output shapes.
CAST provides the prompting foundation: TADA-Agent adapts Algorithmic
Prompting and Thinking-before-Speaking to the two contracts
~\citep{xie2026cast}. Atom-model selection is independent of orchestration,
allowing efficient lightweight models to handle high-volume calls.

\subsection{Skill-Based Implementation}
\label{sec:skill-implementation}

TADA-Agent is implemented as a text-analysis skill. When activated, the skill
provides the two API signatures, input and output conventions, composition
rules, and canonical examples. The skill guides the orchestrator; the
code-execution tool runs the Python workflow in which the APIs are available.
The orchestrator still decides how to compose them with symbolic code.

The public contracts remain unchanged across input scales. Batching, parallel
calls, validation, recovery, independent versus joint Tagging, and category
consolidation remain inside the library. The orchestrator therefore uses one
analytical workflow rather than constructing a separate MapReduce program
(Section~\ref{sec:scalable}).

\section{Large-Scale TADA Execution}
\label{sec:scalable}

Large-scale text analysis follows a MapReduce-style semantic execution rather
than simple chunking and concatenation. Consider discovering the top-$k$
topics from a textual column with thousands of rows. Categorizing row chunks
independently does not produce the global top-$k$: chunks may omit globally
important topics, use synonymous names, or choose different levels of
granularity.

The execution addresses a distinct problem at each stage. The map stage
partitions the input under a chunking configuration fixed by model capacity,
local stability, and consolidation cost. Per-chunk processing must preserve enough
information for the global request, motivating a local candidate budget
$r\geq k$. The reduce stage then semantically consolidates local proposals
into one representative global result. Independent Tagging is the simpler
row-aligned case and can be reassembled by \texttt{rid}; joint Tagging first
uses large-scale Categorization to establish one shared output list.
Partitioning, parallel calls, consolidation, and recovery remain inside the
two atom implementations, leaving their public APIs unchanged.

\subsection{Planning a Large-Scale Request}
\label{sec:large-planning}

The central large-scale problem is to discover the global top-$k$ categories
from $N$ text items that cannot be processed reliably in one call.  TADA
randomly partitions the input into chunks of approximately $n$ rows, asks each
chunk for its top-$r$ category candidates, and consolidates the local
proposals into the final top-$k$.  Planning therefore fixes the chunk size
$n$ under model capacity and selects the local retention budget $r\geq k$.

These parameters are coupled.  Smaller chunks exhibit greater sampling
variation, so a globally important category is more likely to fall below a
local cutoff; retaining a larger $r$ provides a buffer.  Larger chunks reduce
this variation and can use an $r$ closer to $k$.  The category distribution
also matters: if the $k$-th most frequent category is clearly more frequent
than the $(k{+}1)$-th, $r=k$ suffices, whereas several nearly equally frequent
categories around rank $k$ require each chunk to keep more than $k$
candidates.  Increasing $r$
improves preservation but raises mapper output and consolidation cost.

Once $n$ is fixed and $r$ is selected, large-scale Categorization follows the
partition--consolidate plan in Section~\ref{sec:large-categorization}.
Independent Tagging directly partitions and reassembles rows, while Joint
Tagging first uses large-scale Categorization to establish one shared output
domain (Section~\ref{sec:large-tagging}).  The next subsection formalizes the
remaining planning question: given $n$, how to select $r$ so that local
retention preserves the requested global top-$k$.
\subsection{Top-\texorpdfstring{$k$}{k} Preservation}
\label{sec:topk-theory}

We analyze an idealized version of this pipeline. Each text item belongs to
exactly one category. After the items are randomly divided into chunks, each
chunk reports its $r$ most frequent categories and their exact local counts;
the reducer merges identical categories across chunks and returns the $k$
largest combined counts.  Planning is therefore a top-$k$ preservation problem: when
does this local top-$r$ truncation recover the full population's top-$k$?

Formally, let $Y(x)\in\{1,\ldots,C\}$ be the category of item $x$, and order
the full-input supports as
\begin{equation}
    p_i
    =
    \frac{1}{N}
    \sum_{x\in X}\mathbf 1[Y(x)=i],
    \quad
    \Delta_{i,r}=p_i-p_{r+1},
    \label{eq:rank-sensitive-gap}
\end{equation}
with the categories indexed so that $p_1\geq\cdots\geq p_C$.
Let $T_k$ be the full-input top-$k$ and $\widehat T_k$ the result after local
retention.  For $B$ balanced chunks with minimum size $n_{\min}$, the full analysis
(\suppTopkBootstrap) gives the sufficient bound
\begin{equation}
    \Pr\!\left(\widehat T_k=T_k\right)
    \geq
    1-
    B(C-r)
    \sum_{i=1}^{k}
    \exp\!\left(
      -\frac{n_{\min}\Delta_{i,r}^{2}}{2}
    \right).
    \label{eq:topk-preservation-bound}
\end{equation}
The bound makes the trade-off explicit. Increasing chunk size reduces local
sampling variation; increasing $r$ moves the comparison from $p_{k+1}$ to
the smaller $p_{r+1}$ and widens $\Delta_{i,r}$. A large frequency gap after rank $k$
allows $r=k$, whereas nearly equal frequencies around rank $k$ require a
larger $r$, at the price of longer local outputs and more consolidation work.

\paragraph{Operational selection.}
TADA uses a random pilot sample to select $r$.  The sample is sized to fit
within one Categorization call and one Tagging call.  Categorization identifies
the candidate set from the sample, Tagging maps the pilot rows to that list,
and symbolic counting yields category counts $c_1,\ldots,c_C$ and support
estimates $\widehat p_i=c_i/N_{\mathrm p}$.

Rather than substituting these noisy estimates directly into
Equation~\ref{eq:topk-preservation-bound}, TADA propagates their uncertainty
through a paired bootstrap.  For each outer replicate $u$, it draws a support
profile from the Jeffreys Dirichlet posterior induced by the pilot counts.
Conditional on that profile, it simulates $V$ balanced partitions for the
target $N$ and $n$.  For every candidate $r$, it retains the exact local
top-$r$ counts and compares the combined result with the realized full-input
top-$k$.  All candidates share the same support profiles and partitions.  If
$Z_{u,v}(r)$ indicates recovery in partition replicate $v$, then
\begin{equation}
    \widehat\rho_u(r)
    =
    \frac{1}{V}\sum_{v=1}^{V}Z_{u,v}(r).
    \label{eq:conditional-bootstrap-recovery}
\end{equation}
Let $Q_{\alpha}(r)$ be a lower empirical quantile of
$\widehat\rho_u(r)$ across pilot-profile draws.  Given a target recovery
$1-\delta$, TADA selects
\begin{equation}
    \widehat r
    =
    \min\left\{
      r\in\mathcal R:
      Q_{\alpha}(r)\geq 1-\delta
    \right\}.
    \label{eq:bootstrap-r-selection}
\end{equation}
Thus, the planner chooses the smallest tested retention budget that meets the
target. In TADA-Agent, we set the batch size to $n=1{,}000$, which is suitable
for commonly used lightweight LLMs. Appendix~\ref{app:topk-bootstrap}
gives the theorem, proof, bootstrap details, and the assumptions separating
retention error from semantic assignment and consolidation error.
\subsection{Large-Scale Categorization}
\label{sec:large-categorization}
Once $n$ is fixed and $r$ is selected
(Sections~\ref{sec:large-planning}--\ref{sec:topk-theory}),
TADA-Agent randomly shuffles the $N$
input rows and partitions them into
$B=\lceil N/n\rceil$ balanced chunks $X_1,\ldots,X_B$. It constructs a
local query $q_r$ from the user query by requesting up to $r$ categories,
ordered by their prominence in the chunk. Categorization then runs in
parallel:
\begin{equation}
    \mathcal C_b
    =
    \textsc{Categorize}(X_b,q_r),
    \quad
    |\mathcal C_b|\leq r,
    \quad
    b\leq B.
    \label{eq:local-categorization}
\end{equation}

Despite random shuffling, the local lists
$\mathcal C_1,\ldots,\mathcal C_B$ may still use synonymous names or
different descriptions for the same category. TADA-Agent
therefore consolidates names and descriptions across all local lists and
uses the pilot support estimate $\widehat p$ to select the $k$ categories
that best represent the full input:
\begin{equation}
    \widehat{\mathcal C}_k
    =
    \textsc{SelectK}\!\left(
      \textsc{Consolidate}\!\left(
        \mathcal C_1,\ldots,\mathcal C_B;\widehat p,q
      \right),
      k
    \right).
    \label{eq:large-categorization}
\end{equation}
\subsection{Large-Scale Tagging}
\label{sec:large-tagging}

Consider two Tagging requests.  ``Extract the postcode from each address'' is
\emph{independent}: each result depends only on its row and the same query.
``Assign each review one of at most 20 recurring keywords'' is \emph{joint}:
the rows must share one output domain, and that domain cannot be determined
from any row alone; it must first be inferred from the full dataset.

Therefore, Independent Tagging can process chunks directly under the same
query, whereas Joint Tagging requires a global discovery step before parallel
row assignment.  TADA reuses large-scale Categorization for this step; once
the shared domain is fixed, row assignment becomes an Independent Tagging
task.

\paragraph{Independent Tagging.}
TADA-Agent partitions an independent request, runs Tagging in parallel, and
reassembles the row-aligned strings by stable identifier:
\begin{align}
    S_b
    &=
    \textsc{Tag}(X_b,q),
    \qquad b=1,\ldots,B,
    \label{eq:independent-tagging}\\
    S
    &=
    \textsc{ReassembleByRid}(S_1,\ldots,S_B).
    \label{eq:tagging-reassembly}
\end{align}

\paragraph{Joint Tagging.}
TADA-Agent first derives a Categorization query $q_C$ that discovers the
shared domain, then constructs an assignment query $q_A$ and uses the
independent path:
\begin{align}
    \mathcal C
    &=
    \textsc{Categorize}_{\mathrm{large}}(X,q_C),
    \label{eq:joint-domain}\\
    q_A
    &=
    \textsc{BuildAssignmentQuery}(q,\mathcal C),\\
    S
    &=
    \textsc{Tag}_{\mathrm{ind,large}}(X,q_A).
    \label{eq:large-joint-tagging}
\end{align}
For the keyword example, $q_C$ asks Categorization to identify up to 20
representative keyword labels shared across the dataset.  The assignment
query $q_A$ lists their names and descriptions and asks each row to return its
most relevant keyword.

\section{Evaluation}
\label{sec:evaluation}

We organize the evaluation around three questions:
\begin{description}
  \item[RQ1: Quality, reliability, and request cost.] Across the TABench scenario taxonomy, how does TADA-Agent compare with a semantic-operator system and non-semantic baselines in row-level quality, complete-artifact reliability, and dedicated semantic-model requests?
  \item[RQ2: Large-scale execution.] Beyond one reliable context window, does
  batched Tagging preserve assignments, and does Categorization retain the
  candidates needed for the global top-$k$?
  \item[RQ3: Transfer beyond TABench.] Does the Tagging interface, unchanged, transfer to established legal and biomedical extraction benchmarks?
\end{description}

\subsection{RQ1: Quality, Artifact Reliability, and Request Cost}
\label{sec:eval-rq1}

\begin{table*}[t]
\caption{Row-level semantic quality on TABench by scenario class.
Except for the semantic-ranking row, each entry is the percentage of
evaluated units judged at least as good as the common reference output, with a
95\% case--run cluster-bootstrap interval over three independent runs per
configuration (identical inputs, models, and judge). The semantic-ranking
row reports mean within-case Spearman $\rho$ against the same reference,
macro-averaged over the four ranking cases and three runs, with a case--run
cluster-bootstrap 95\% interval computed the same way (wide because only
four cases enter this row); the complementary absolute-score agreement is
reported in the text, and Direct is omitted because its
incomplete/malformed Q34 artifacts yield method-dependent row coverage.
Bold marks the highest point estimate and does not imply statistical
significance; classes are never aggregated.}
\label{tab:rq1-per-group}

\centering
\footnotesize
\setlength{\tabcolsep}{4.5pt}
\begin{tabular}{lccccc}
\toprule
Scenario class
& TADA-Agent
& Semantic operators
& Claude Code
& Code-only
& Direct \\
\midrule
Classification
& $\mathbf{98.6}$ {\scriptsize[97.0, 100.0]}
& $97.5$ {\scriptsize[96.1, 98.8]}
& $97.9$ {\scriptsize[95.2, 100.0]}
& $96.6$ {\scriptsize[94.0, 98.8]}
& $94.7$ {\scriptsize[91.3, 98.0]} \\

Field extraction
& $\mathbf{98.9}$ {\scriptsize[96.6, 99.9]}
& $98.2$ {\scriptsize[95.5, 99.7]}
& $97.9$ {\scriptsize[93.9, 99.7]}
& $90.6$ {\scriptsize[80.3, 98.4]}
& $82.5$ {\scriptsize[72.2, 95.0]} \\

Semantic ranking (Spearman $\rho$)
& $\mathbf{0.801}$ {\scriptsize[0.600, 0.980]}
& $0.752$ {\scriptsize[0.483, 0.957]}
& $0.774$ {\scriptsize[0.558, 0.962]}
& $0.590$ {\scriptsize[0.249, 0.921]}
& --- \\

Constrained generation
& $\mathbf{89.1}$ {\scriptsize[83.8, 94.0]}
& $78.6$ {\scriptsize[64.2, 88.8]}
& $84.2$ {\scriptsize[74.8, 90.6]}
& $35.0$ {\scriptsize[27.5, 42.8]}
& $60.3$ {\scriptsize[55.2, 64.0]} \\
\bottomrule
\end{tabular}
\end{table*}

\paragraph{Benchmark and protocol.}
TABench evaluates 40 complete analysis requests over spreadsheet tables
(workbooks) rather than isolated semantic calls. Its taxonomy derives from a
structured review and contains 12 scenario classes organized along analytical
scenario, semantic operation, and input--output structure (construction
details in Appendix~\ref{app:tabench-construction}). Each case provides a
source table, a natural-language request, an output contract, and evaluation
criteria. The 40 scored cases cover ten scenario classes with three to five
cases each (Table~\ref{tab:tabench-inventory-classes}). Their source tables
have 89--201 rows and 2--8 columns (200 rows in 38 cases), 7{,}890 rows in
total. Seven cases carry rule-based answers derived
from generator-injected phenomena such as near-duplicates and entity aliases.

All configurations receive the same inputs and requests. Because most TABench
requests have no unique answer, we use one held-out TADA run as a common
reference output rather than assign absolute scores. This reference is fixed
for every method, including three independent TADA repetitions. Row-aligned
outputs without rule-based answers are judged per differing cell
(Table~\ref{tab:rq1-per-group}); other non-deterministic outputs are judged as
complete artifacts (Table~\ref{tab:rq1-artifact}). Seven cases with rule-based
answers instead use absolute accuracy and enter neither table. For
reference-based comparisons, an anonymized, position-swapped LLM judge casts
three votes in each A/B order; a unit wins only when both orders give the same
majority result.
If the two orders disagree, the comparison is a tie. On integer-rating tasks,
scores that differ by one scale step are also treated as ties. Coverage is
audited over all 40 cases; a missing required row or output is a coverage
failure rather than a successful judgment unit.

\paragraph{Controlled configurations.}
The controlled configurations use the same public frontier models where their
execution paths permit:
Claude Opus 5 hosts TADA-Agent and the Claude Code baseline; semantic calls
(atom or operator) execute on Claude Sonnet 5, and GPT-5.5 serves as the judge. The five
configurations share the frozen tasks and evaluator but differ in the semantic
interface and execution support available to the host agent: (i)~the full
system with both semantic atoms; (ii)~\emph{semantic operators}, an external
semantic-operator system executed through its official Python APIs on the
same semantic model and fixed serving endpoint, instantiated with
LOTUS~1.2.4~\citep{patel2025lotus};
(iii)~\emph{Claude Code}, run without the TADA skill or atom APIs, so the coding agent must choose its own decomposition; (iv)~\emph{Code-only}, which must express all semantic judgments as symbolic code; and
(v)~\emph{Direct}, a single prompted call without agent scaffolding.
Each configuration is run three times independently to expose
run-to-run variance.

The three measurements are reported separately. The seven rule-based cases are
at or near ceiling for both TADA-Agent and semantic operators (summary results
appear in the appendix). Confidence intervals use
case--run cluster bootstrap for row units and descriptive Wilson intervals for
artifacts. We use 10{,}000 bootstrap resamples with a fixed analysis seed;
all task inputs are held fixed, while the three independent model executions
capture run-to-run stochasticity. Scenario placement uses method-independent
criteria: a case is reported separately when its output contract or available
metric does not measure the named scenario. The normalization-and-repair class
therefore appears in the appendix; its only row-level case is at ceiling
(99.8--100.0 across methods), and its remaining cases are deterministic
contract checks.

\paragraph{Semantic quality.}
Table~\ref{tab:rq1-per-group} shows a mixed result. Classification and field
extraction are near ceiling for both systems: TADA-Agent differs from semantic
operators by 1.1 and 0.7 points. These classes mainly test whether the caller
can apply a well-defined semantic decision consistently across rows, so the
two interfaces offer similar quality when the desired output is already
row-aligned.

The distinction is clearer on ranking and constrained generation. For
ranking, Table~\ref{tab:rq1-per-group} reports mean within-case Spearman
correlation with the reference output: TADA reaches 0.801 versus 0.752 for semantic
operators, though the four-case intervals overlap widely. A complementary
absolute-score agreement measure favors semantic operators (86.6\% versus
81.9\%); it is the percentage of scored items that exactly match the
reference rating or differ by one scale step, macro-averaged over case--run
units. Unlike Spearman $\rho$, it rewards reproducing the reference's numerical
scale as well as its ordering.
Semantic operators better match the reference's scale calibration, while
TADA-Agent's ordering agrees more with the reference. Because the reference is
itself a TADA run, this ordering agreement measures TADA-Agent's run-to-run
consistency but measures the baselines' agreement with a different system; it
is not evidence of better ranking quality. Both systems order the standard ranking cases
nearly identically ($\rho\geq0.93$), and most of the difference comes from one
calibration-sensitive scoring case. On constrained generation TADA leads
(89.1 versus 78.6); the advantage
arises when a semantic value must subsequently satisfy a workbook-level
cardinality, distribution, or formatting rule that symbolic code can enforce
exactly. We do not interpret any overlapping intervals as significant
differences.

\paragraph{Artifact reliability and request cost.}
The broader difference appears in whether each system delivers a complete
artifact without omitting required content (Table~\ref{tab:rq1-artifact}).
Relative to the same fixed reference, TADA-Agent has higher preservation point
estimates than semantic operators in all five classes. The largest
observed gaps occur in grouping (8/9 versus 5/9), aggregation and summarization
(12/12 versus 6/12), and anomaly detection (6/6 versus 4/6). Filtering is
closer (9/12 versus 8/12), while numerical and temporal analysis remains
uncertain at 6/9 versus 4/9. The largest differences occur when several judgments and symbolic steps must
survive into one delivered workbook. Because the denominators are small, these
estimates are descriptive and specific to this reference; we do not aggregate
them into a single TABench score. They are better read as evidence about
failure modes than as a ranking of systems. Claude Code without the atoms
matches TADA-Agent's estimate in three of the five classes, and single-call
Direct prompting matches it in two. The largest gap to semantic operators is
in aggregation and summarization, where semantic operators also score below
Direct prompting; we have not analyzed the causes of these failures.

\begin{table*}[t]
\caption{Complete-artifact preservation on TABench. Each entry counts outputs
judged at least as complete and correct as the fixed reference, with descriptive
95\% Wilson intervals in percent. The unit is a delivered artifact rather than
a row; functionally determined identity columns do not count as missing content.
These results are specific to this reference output and are not combined with
Table~\ref{tab:rq1-per-group}.}
\label{tab:rq1-artifact}
\centering
\footnotesize
\setlength{\tabcolsep}{3.5pt}
\begin{tabular}{lccccc}
\toprule
Scenario class
& TADA-Agent
& Semantic operators
& Claude Code
& Code-only
& Direct \\
\midrule
Grouping and clustering
& $\mathbf{8/9}$ {\scriptsize[56.5, 98.0]}
& $5/9$ {\scriptsize[26.7, 81.1]}
& $\mathbf{8/9}$ {\scriptsize[56.5, 98.0]}
& $6/9$ {\scriptsize[35.4, 87.9]}
& $5/9$ {\scriptsize[26.7, 81.1]} \\
Semantic filtering
& $\mathbf{9/12}$ {\scriptsize[46.8, 91.1]}
& $8/12$ {\scriptsize[39.1, 86.2]}
& $6/12$ {\scriptsize[25.4, 74.6]}
& $4/12$ {\scriptsize[13.8, 60.9]}
& $0/12$ {\scriptsize[0.0, 24.2]} \\
Aggregation and summarization
& $\mathbf{12/12}$ {\scriptsize[75.8, 100.0]}
& $6/12$ {\scriptsize[25.4, 74.6]}
& $\mathbf{12/12}$ {\scriptsize[75.8, 100.0]}
& $11/12$ {\scriptsize[64.6, 98.5]}
& $\mathbf{12/12}$ {\scriptsize[75.8, 100.0]} \\
Numerical and temporal analysis
& $\mathbf{6/9}$ {\scriptsize[35.4, 87.9]}
& $4/9$ {\scriptsize[18.9, 73.3]}
& $\mathbf{6/9}$ {\scriptsize[35.4, 87.9]}
& $\mathbf{6/9}$ {\scriptsize[35.4, 87.9]}
& $1/9$ {\scriptsize[2.0, 43.5]} \\
Anomaly detection
& $\mathbf{6/6}$ {\scriptsize[61.0, 100.0]}
& $4/6$ {\scriptsize[30.0, 90.3]}
& $3/6$ {\scriptsize[18.8, 81.2]}
& $5/6$ {\scriptsize[43.6, 97.0]}
& $\mathbf{6/6}$ {\scriptsize[61.0, 100.0]} \\
\bottomrule
\end{tabular}
\end{table*}

Request cost separates the systems more sharply. Across 120 task-runs, TADA
makes 451 successful dedicated semantic-model requests (3.76 per task-run),
counting intent detection, data batches, and category consolidation. The
recorded semantic-operator executions contain at least 18{,}852 requests
($\geq157.10$ per task-run), at least 41.8 times the TADA count under the
evaluated native execution paths. This comparison describes the physical
request fan-out of the two implementations; it does not imply that every
operator interface must execute per row. Host-agent
and judge calls are excluded on both sides; the metric counts requests sent to
the dedicated semantic-processing model. The LOTUS total is a strict lower
bound because failed or unpersisted requests cannot always be reconstructed.

\begin{table}[t]
\caption{Recorded calls to the dedicated semantic-processing LLM across three
runs of each TABench task. ``Mean per task-run'' is the mean number of such LLM
calls, not elapsed time. The $\geq$ symbol marks conservatively incomplete
request logs.}
\label{tab:rq1-cost}
\centering
\footnotesize
\begin{tabular}{lcc}
\toprule
Metric & TADA-Agent & Sem.\ operators \\
\midrule
Task-runs & 120 & 120 \\
Successful requests & \textbf{451} & $\geq$18{,}852 \\
Mean calls per task-run & \textbf{3.76} & $\geq$157.10 \\
Relative count & \textbf{1.0$\times$} & $\geq$41.8$\times$ \\
\bottomrule
\end{tabular}
\end{table}

\paragraph{Validity and error analysis.}
Direct prompting drops 461--643 rows per run on row-aligned tasks, so its
scores reflect incomplete delivery as well as weaker judgments. Code-only
preserves coverage but loses semantic quality in all three runs. TADA
satisfies every deterministic output contract, including required sheets and
columns, row coverage, length limits, and required scoring scales, with zero coverage
failures. These baselines separate two roles: agent scaffolding supports
completion, while the atoms supply reproducible semantic judgments.

Evaluator notes show how the alternatives fail. Lexicon-based sentiment
mislabels sarcastic feedback, and TF--IDF clustering groups shared surface
terms into incoherent issue categories. Other runs complete the semantic step
but omit required filtering, aggregation, or workbook mutation. The rubric
does not penalize an alternative method when it produces the correct workbook;
these cases instead motivate evaluating delivered artifacts rather than tool
invocation or a plausible chat explanation.

Judge agreement is high: unanimous three-vote agreement ranges from 88.9\%
to 94\%. Six human annotators independently label a stratified sample of 39
units under the same anonymized A/B presentation; the sample oversamples
contested and low-confidence judgments. Of the 33 units with a human majority,
the judge matches 26 (78.8\%, Cohen's $\kappa=0.68$), including 17/19 in the
contested stratum, with two direction reversals. Its mean agreement with
individual annotators (0.60) exceeds their mean pairwise agreement (0.55;
Fleiss $\kappa=0.32$). The six units without a human majority are adjacent ratings on the reference
scale, which the deterministic rule treats as ties.

Within the scope of this comparison, TADA retains competitive row-level
quality with higher complete-artifact preservation estimates, full contract
compliance, and far fewer semantic-model requests, without uniformly better
isolated judgments. The comparison is between complete systems. It does not
separate the atom interface from TADA-Agent's prompting, batching, and skill
guidance, and it does not measure how often baseline plans recompute a
semantic variable. We therefore read these results as evidence that the
interface is workable and efficient in practice, not as a causal attribution
of the gains to the interface alone.

\subsection{RQ2: Large-Scale Execution}
\label{sec:eval-rq2}

RQ2 evaluates end-to-end Tagging and isolates one scaling component of
Categorization. The first experiment tests whether
batched Tagging preserves row assignments beyond one reliable context window.
The second tests whether locally retained candidates recover the global
top-$k$ under fixed canonical assignments.

\paragraph{End-to-end scaling.}
We use three fixed-taxonomy workloads: four-way sentiment and topic assignment
over 32,179 Play Store reviews, and category assignment over 6,628 synthetic
enterprise-feedback messages. Three strong-model annotation rounds produce
reference labels by majority vote; rows without a two-vote majority
(0.2--3.1\%) are excluded. TADA-Agent uses the internal Tagging path with
$n=1{,}000$ for three repetitions per workload. Claude Code runs under a
controlled coding-agent protocol: each repetition starts in a fresh directory containing
only the data file and query, with configuration, skills, and tool extensions
disabled; the agent chooses its own method and decomposition. LOTUS
\texttt{sem\_map} uses its standard per-row projection. Its optimizer covers
filters, joins, top-$k$, and group-by, but not semantic maps, so one call per
row is the native execution path for this workload. Because repeated 32K
runs exceed our serving budget, Table~\ref{tab:rq2-e2e} marks its single
measured subset results ($500$ rows for the 32K workloads and $2{,}050$
rows from a checkpointed enterprise-feedback run); quality is not
extrapolated beyond what was measured.

We report accuracy, macro-F1, and accuracy normalized by mean pairwise
agreement among the three annotation rounds. Macro-F1 averages per-class F1,
so minority classes receive equal weight. A normalized score of 1.0 means
agreement with the voted reference matches inter-annotator agreement; it is a
calibration point, not an upper bound. Pairwise agreement is
0.957/0.908/0.762 across the three workloads.

\paragraph{End-to-end results.}
TADA-Agent has the highest observed full-scale accuracy on all three workloads
(0.973, 0.942, and 0.837), with normalized accuracy at or above 1.0. It is also
stable across the three repetitions: the accuracy standard deviations are
0.000, 0.002, and 0.015. Claude Code varies more across runs. On topic
assignment, its accuracy ranges from 0.723 to 0.928
($\mathrm{s.d.}=0.116$); on sentiment, its macro-F1 is 0.577 versus 0.928 for
TADA, showing that most errors occur in minority classes. One earlier Claude
Code run timed out after three hours without producing an output and was rerun
under the same controlled protocol. LOTUS falls between the two full-scale
systems in normalized accuracy (0.93--1.00) on the measured subsets, but these
are not matched full-workload repetitions.

\paragraph{Execution cost.}
TADA processes one batch per semantic request: 33 requests for each 32K-row run
and 7 for each 6.6K-row run. For the two 32K workloads, a clean run uses
1.2--1.6M provider-reported tokens and takes 12--37 minutes; the 6.6K workload
uses 0.29--0.34M tokens and completes in 2--10 minutes. One sentiment run encountered transient provider
failures, increasing observed request events to 105 and wall time to 165 minutes
without changing the planned 33-request computation. LOTUS instead invokes its
native semantic map once per row. A 32K run therefore requires 32{,}179 calls,
or $975\times$ TADA's planned request count. Extrapolating from measured
per-row execution gives 4--6M tokens and 36--42 hours at the same concurrency.
Claude Code uses 22--155 API messages per completed run, with token accounting
dominated by cache reads. Full telemetry and metering caveats appear in
\suppEECost.

\begin{table*}[t]
\caption{Large-scale end-to-end execution. Accuracy is agreement with the
majority-vote labels. \emph{Acc./agreement} divides accuracy by inter-annotator
agreement (1.0 means they match). TADA-Agent and Claude Code report mean
$\pm$ s.d. over three full-scale runs; LOTUS rows are measured subsets:
500 rows for the 32K workloads and the first 2{,}050 rows for the 6.6K
workload ($\dagger$). Macro-F1 and Acc./agreement are also means across valid
full-scale repetitions; only Accuracy displays its across-run s.d.}
\label{tab:rq2-e2e}
\centering
\footnotesize
\begin{tabular}{llccc}
\toprule
Workload & System & Accuracy & Macro-F1 & Acc./agreement \\
\midrule
\multirow{3}{*}{\shortstack[l]{Sentiment\\32{,}179 rows}}
 & TADA-Agent & $\mathbf{0.973{\scriptstyle\pm0.000}}$ & $\mathbf{0.928}$ & $\mathbf{1.017}$ \\
 & Claude Code & $0.817{\scriptstyle\pm0.047}$ & $0.577$ & $0.854$ \\
 & LOTUS$^\dagger$ & $0.954$ & $0.884$ & $0.997$ \\
\midrule
\multirow{3}{*}{\shortstack[l]{Topic assign.\\32{,}179 rows}}
 & TADA-Agent & $\mathbf{0.942{\scriptstyle\pm0.002}}$ & $\mathbf{0.838}$ & $\mathbf{1.038}$ \\
 & Claude Code & $0.856{\scriptstyle\pm0.116}$ & $0.729$ & $0.943$ \\
 & LOTUS$^\dagger$ & $0.881$ & $0.838$ & $0.970$ \\
\midrule
\multirow{3}{*}{\shortstack[l]{Category assign.\\6{,}628 rows}}
 & TADA-Agent & $\mathbf{0.837{\scriptstyle\pm0.015}}$ & $\mathbf{0.829}$ & $\mathbf{1.098}$ \\
 & Claude Code & $0.658{\scriptstyle\pm0.013}$ & $0.605$ & $0.863$ \\
 & LOTUS$^\dagger$ & $0.706$ & $0.671$ & $0.926$ \\
\bottomrule
\end{tabular}
\end{table*}

\paragraph{When categories must also be discovered.}
The preceding experiment supplies the target taxonomy to every system. We also
test a harder variant in which Claude Code must first invent a category
vocabulary and then assign rows to it, matching the role of TADA's joint
Tagging path (Section~\ref{sec:large-tagging}). After aligning its discovered
labels to the reference taxonomy, Claude Code reaches $0.68 \pm 0.01$ accuracy
on the 32K topic workload and $0.21 \pm 0.08$ on enterprise feedback. In each
run, 3--8 discovered categories have no reference counterpart, mainly because
the agent merges categories at a different granularity or invents catch-all
buckets. The alignment protocol and discovered vocabularies appear in
Appendix~\ref{app:e2e-openvocab}.

\paragraph{Large-scale Categorization.}
The controlled study validates the candidate-retention stage of large-scale
Categorization, isolated from semantic assignment and
consolidation. We freeze six $N=32{,}000$ populations over 20 canonical
categories, fix $k=5$ and $n=1{,}000$, and evaluate $r=5,\ldots,20$. One
population has a clear cutoff. The other five share three near-cutoff
competitors and a geometric tail, with only the cutoff gap narrowing across
levels (generation details in \suppRQControlled{}).

A 1,000-row pilot drives the selection rule from
Section~\ref{sec:topk-theory}. To account for sampling uncertainty, we draw 100
plausible category-frequency profiles from the pilot and simulate 20 matched
random partitions for each profile. For evaluation, each candidate $r$
is tested over 5,000 fresh random partitions under the known full
distribution, with no LLM call during this test. This separates planning under
pilot uncertainty from evaluation on the held-out full population.

\begin{figure}[t]
\centering
\includegraphics[width=\columnwidth]{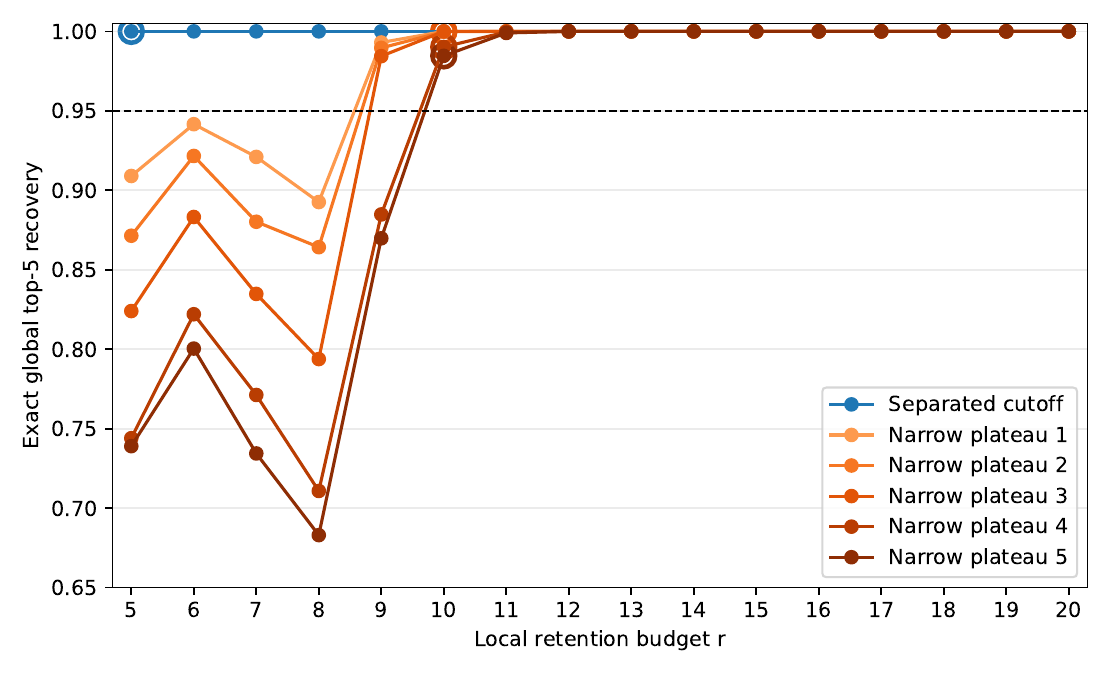}
\caption{Probability of recovering the true global top-5 categories over
5,000 random partitions of each
fixed 32K-row population as the local retention budget varies from
$r=5$ to 20. The dashed line is the 0.95 target; outlined circles mark
pilot-selected values. The blue separated distribution needs no extra
candidates; the five orange curves progressively narrow the cutoff gap.}
\Description{Recovery curves for six fixed category distributions as the local
retention budget increases. The separated distribution remains at full
recovery, while near-tied distributions generally improve with larger budgets.}
\label{fig:rq2-bootstrap}
\end{figure}

Figure~\ref{fig:rq2-bootstrap} shows that every $r$ recovers the separated
top-5 and the planner selects $r=5$. Across the distributions with small
frequency gaps, fixed-$r=5$
recovery falls from 90.9\% to 73.9\%. The smallest $r$ whose held-out recovery
reaches 0.95 is 9 for levels 1--3 and 10 for levels 4--5; the pilot rule selects
$r=10$ throughout. Thus it retains one extra category in the first three levels
and selects the minimum sufficient value in the last two. Recovery at the
selected budget is 98.5--100\%.
The analytical bound describes how chunk size, cutoff gap, and tail width
control retention; the bootstrap gives the operating estimate under sampling
and partition uncertainty.

\subsection{RQ3: Transfer Beyond TABench}
\label{sec:eval-rq3}

To test whether the same interface applies beyond TABench, we reproduce two
public document-extraction benchmarks used in related work, retaining their
native tasks and evaluators. CUAD~\citep{hendrycks2021cuad} tests legal clause
extraction, and BioDEX~\citep{doosterlinck2023biodex} tests biomedical adverse-
event extraction. Both can be expressed through the unchanged Tagging
interface, which lets this experiment test cross-domain transfer without adding
domain-specific atoms; benchmark-specific postprocessing remains outside the
atom.

\begin{figure}[t]
  \centering
  \includegraphics[width=0.86\columnwidth,keepaspectratio]{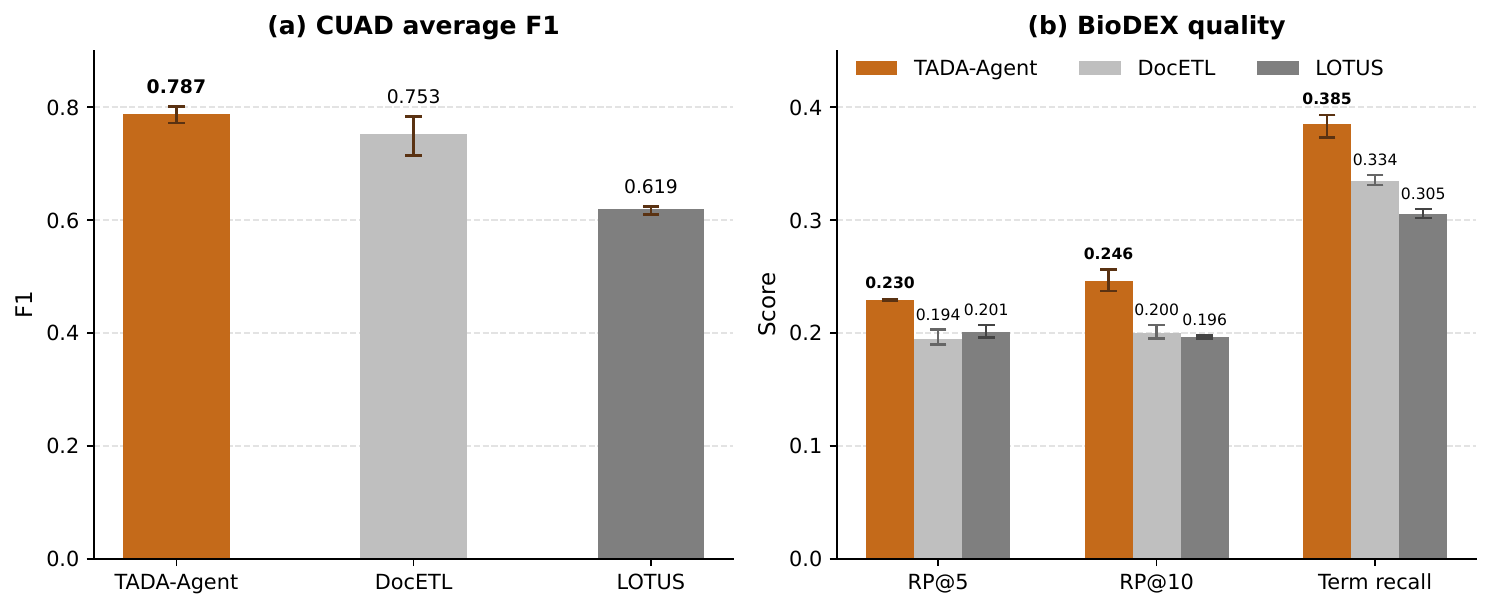}
\caption{Public document-extraction benchmarks (higher is better).
\textbf{(a)} Reproduced CUAD average F1; TADA-Agent, DocETL, and LOTUS
rerun with the same \texttt{GPT-5.5} model and evaluator. \textbf{(b)}
Reproduced BioDEX report precision at 5 and 10 predictions (RP@5/10) and
recall of gold adverse-event terms under the same model,
candidate budget, and official evaluator; bars are means over completed
runs, and error bars show the observed min--max where repeats exist.
Provenance caveats are stated in the text.}
\Description{Two grouped bar charts compare TADA-Agent, DocETL, and LOTUS on
CUAD average F1 and BioDEX RP at 5, RP at 10, and term recall.}
  \label{fig:rq3-public-benchmarks}
\end{figure}

CUAD evaluates query-specific legal clauses as row-aligned Tagging followed by
deterministic normalization. BioDEX contains 100 drug-safety reports evaluated
against 24,313 MedDRA terms. TADA-Agent Tags adverse-reaction phrases and then uses benchmark-specific
retrieval and deterministic alignment against a top-200 MedDRA candidate set.
Three completed TADA runs use the same model and candidate budget. The DocETL
and LOTUS baselines are single protocol-matched runs; unmatched repetitions are
excluded. Figure~\ref{fig:rq3-public-benchmarks} therefore shows uncertainty
only where comparable repeats exist.

On CUAD, TADA-Agent reaches 0.787 average F1, compared with 0.753 for DocETL
and 0.619 for LOTUS under the common environment. On BioDEX, its three-run
means are 0.230 RP@5, 0.246 RP@10, and 0.385 term recall, with standard
deviations 0.0009, 0.0095, and 0.0104. The recorded LOTUS point is
(0.194, 0.200, 0.334) and DocETL is (0.201, 0.196, 0.305). Thus the same
Tagging interface produces the highest observed CUAD score and higher BioDEX
point estimates under the reproduced protocols, subject to the uneven
baseline repetition.

\section{Discussion, Limitations, and Conclusion}
\label{sec:conclusion}

TADA places the semantic interface at reusable state: Tagging materializes
one string per input row, Categorization a category table, and exact
operations remain in the symbolic runtime. This is an interface choice, not an
increase in expressive power. A semantic-operator system that offers a
semantic map, or exposes intermediate decisions, can express the same
workflows, and a careful plan over such operators can avoid the repeated
computation shown in Figure~\ref{fig:interface-comparison}(a). TADA makes
that plan the default rather than an option the agent must find. Composite semantic operators
remain appropriate when their final artifact completes the task. TADA targets
multi-step analyses in which later operations must reuse semantic results.
The choice is therefore workload-dependent. A one-shot filter or top-$k$
request may benefit from an optimized composite implementation; a workflow
that later groups, models, audits, or revises the same semantic variable
benefits from materializing that variable once.

The evaluation reflects this distinction. Semantic operators remain a strong
baseline for isolated judgments and better reproduce the reference output's
absolute ranking scale. At the workflow level, TADA has higher observed
complete-artifact preservation, satisfies deterministic output requirements,
and invokes the semantic model fewer times. The ranking intervals overlap and
the artifact samples are small; within those bounds, the results show that
reusable state can preserve competitive semantic quality while helping an
agent complete multi-step analytical requests.

Several limitations define the current interface. Tagging returns one string
per input item, and Categorization returns category objects over a text column
or row group. Pairwise semantic joins, global matching, graph-structured
outputs, and comparison-based ranking may require additional atoms when their
outputs cannot be represented by these contracts. Such an extension should
be driven by a recurring output structure that later steps need to reuse, not
by adding one operator for every analytical verb. Richer values also raise
schema questions: confidence, provenance, multiple labels, and structured
records may need explicit contracts rather than serialization into strings.

Large-scale execution introduces a second boundary for both atoms. Independent Tagging can
partition rows without changing the task, whereas joint Tagging first needs a
shared domain. Categorization also depends on retaining locally proposed
categories that may matter globally. Our study isolates this retention stage
under exact assignments, counts, and consolidation. Real deployments add
errors at all three steps, so the bootstrap estimates only recovery under pilot
and partition noise. The tested scales do not establish asymptotic cost or
latency across providers.

Dedicated semantic-model requests compare interface fan-out under a fixed
model and endpoint, not total systems cost. Batches may contain more tokens,
providers may cache or schedule differently, and host-agent reasoning also
costs resources. We therefore report requests separately from tokens and time;
the preferred granularity depends on overhead, rate limits, latency, retries,
and incremental-output needs. In particular, the measured request reduction
compares the evaluated native implementations and does not imply that a
semantic operator cannot itself batch or materialize outputs.

Materialized semantic state also changes how errors propagate. Reuse prevents
a workflow from recomputing the same notion inconsistently, but it can carry
one mistaken value into several later operations. Because the state is stored
as ordinary table data, an agent or user can inspect, revise, and recompute
its downstream effects without rerunning unrelated semantic steps. The present
evaluation measures final artifacts and hard contracts, not the effort needed
to repair an intermediate error; studying this cost is future work.

The logical interface does not prescribe one physical plan: implementations may
cache materialized columns, fuse compatible calls, or consolidate categories
incrementally while preserving row identity and output contracts. Future work
should also study additional contracts for recurring structures, such as
pairwise relations or graphs, that cannot be represented by the current two.

The frozen TABench suite contains 43 requests. The current two-atom interface
supports and scores 40 of them across ten of the 12 reporting classes; the
three boundary cases require text-to-table structuring, semantic imputation,
or semantic joins outside its output contracts. This coverage result motivates
additional atoms for recurring reusable structures rather than one operator per
analytical verb. TABench still does not cover every domain or interaction.
Its row-level and complete-artifact measures depend on a fixed reference and
calibrated LLM judge; human agreement supports but does not remove this
dependence. Because the reference is a held-out TADA run, the ranking
correlation measures TADA's run-to-run consistency but measures the baselines'
agreement with a different system. The evaluation also compares complete
systems: it does not include an ablation that holds prompting and batching
fixed while swapping the atom interface for semantic operators, and it does not
directly count recomputed or inconsistent semantic variables in baseline
plans. End-to-end large-scale Categorization with open category discovery is
likewise not evaluated for TADA; the retention study isolates only the
candidate-retention stage. The public benchmarks broaden the evidence for Tagging, although
BioDEX repetition is uneven. Future work should add matched repetitions and
reference-free or human-authored evaluation targets.

TADA illustrates how domain-specific agents can expose expert-defined semantic
contracts while retaining flexible, query-driven workflow composition.

\bibliography{references}

@inproceedings{yao2023react,
  title = {{ReAct}: Synergizing Reasoning and Acting in Language Models},
  author = {Yao, Shunyu and Zhao, Jeffrey and Yu, Dian and Du, Nan and Shafran, Izhak and Narasimhan, Karthik and Cao, Yuan},
  booktitle = {International Conference on Learning Representations},
  year = {2023},
  url = {https://openreview.net/forum?id=WE_vluYUL-X}
}

@article{patel2025lotus,
  title = {Semantic Operators and Their Optimization: Enabling LLM-Based Data Processing with Accuracy Guarantees in {LOTUS}},
  author = {Patel, Liana and Jha, Siddharth and Pan, Melissa and Gupta, Harshit and Asawa, Parth and Guestrin, Carlos and Zaharia, Matei},
  journal = {Proceedings of the VLDB Endowment},
  volume = {18},
  number = {11},
  pages = {4171--4184},
  year = {2025},
  doi = {10.14778/3749646.3749685},
  url = {https://www.vldb.org/pvldb/vol18/p4171-patel.pdf}
}

@article{hendrycks2021cuad,
  title = {{CUAD}: An Expert-Annotated {NLP} Dataset for Legal Contract Review},
  author = {Hendrycks, Dan and Burns, Collin and Chen, Anya and Ball, Spencer},
  journal = {Advances in Neural Information Processing Systems (NeurIPS) Datasets and Benchmarks Track},
  year = {2021},
  url = {https://arxiv.org/abs/2103.06268}
}

@inproceedings{doosterlinck2023biodex,
  title = {{B}io{DEX}: Large-Scale Biomedical Adverse Drug Event Extraction for Real-World Pharmacovigilance},
  author = {D'Oosterlinck, Karel and Remy, Fran{\c{c}}ois and Deleu, Johannes and Demeester, Thomas and Develder, Chris and Zaporojets, Klim and Ghodsi, Aneiss and Ellershaw, Simon and Collins, Jack and Potts, Christopher},
  editor = {Bouamor, Houda and Pino, Juan and Bali, Kalika},
  booktitle = {Findings of the Association for Computational Linguistics: EMNLP 2023},
  month = dec,
  address = {Singapore},
  publisher = {Association for Computational Linguistics},
  pages = {13425--13454},
  year = {2023},
  doi = {10.18653/v1/2023.findings-emnlp.896},
  url = {https://aclanthology.org/2023.findings-emnlp.896/}
}

@inproceedings{xie2026cast,
  title = {{CAST}: Achieving Stable {LLM}-based Text Analysis for Data Analytics},
  author = {Xie, Jinxiang and Li, Zihao and He, Wei and Ding, Rui and Han, Shi and Zhang, Dongmei},
  booktitle = {Findings of the Association for Computational Linguistics: ACL 2026},
  pages = {2401--2420},
  year = {2026},
  doi = {10.18653/v1/2026.findings-acl.113},
  url = {https://aclanthology.org/2026.findings-acl.113}
}

@article{arora2023evaporate,
  title = {Language Models Enable Simple Systems for Generating Structured Views of Heterogeneous Data Lakes},
  author = {Arora, Simran and Yang, Brandon and Eyuboglu, Sabri and Narayan, Avanika and Hojel, Andrew and Trummer, Immanuel and R{\'e}, Christopher},
  journal = {Proceedings of the VLDB Endowment},
  volume = {17},
  number = {2},
  pages = {92--105},
  year = {2023},
  doi = {10.14778/3626292.3626294},
  url = {https://doi.org/10.14778/3626292.3626294}
}

@article{biswal2024tag,
  title = {Text2SQL is Not Enough: Unifying {AI} and Databases with {TAG}},
  author = {Biswal, Asim and Patel, Liana and Jha, Siddharth and Kamsetty, Amog and Liu, Shu and Gonzalez, Joseph E. and Guestrin, Carlos and Zaharia, Matei},
  journal = {arXiv preprint arXiv:2408.14717},
  year = {2024},
  doi = {10.48550/arXiv.2408.14717},
  url = {https://arxiv.org/abs/2408.14717}
}

@article{codd1970relational,
  title = {A Relational Model of Data for Large Shared Data Banks},
  author = {Codd, Edgar F.},
  journal = {Communications of the ACM},
  volume = {13},
  number = {6},
  pages = {377--387},
  year = {1970},
  doi = {10.1145/362384.362685},
  url = {https://doi.org/10.1145/362384.362685}
}

@inproceedings{dai2024uqe,
  title = {{UQE}: A Query Engine for Unstructured Databases},
  author = {Dai, Hanjun and Wang, Bethany Yixin and Wan, Xingchen and Dai, Bo and Yang, Sherry and Nova, Azade and Yin, Pengcheng and Phothilimthana, Phitchaya Mangpo and Sutton, Charles and Schuurmans, Dale},
  booktitle = {Advances in Neural Information Processing Systems},
  volume = {37},
  pages = {29807--29838},
  year = {2024}
}

@misc{databricks2026aifunctions,
  title = {Enrich Data Using {AI} Functions},
  author = {{Databricks}},
  howpublished = {Databricks Documentation},
  year = {2026},
  note = {Retrieved June 9, 2026},
  url = {https://docs.databricks.com/aws/en/large-language-models/ai-functions}
}

@misc{doris2026aifilter,
  title = {{AI\_FILTER}},
  author = {{Apache Doris}},
  howpublished = {Apache Doris Documentation},
  year = {2026},
  note = {Retrieved June 9, 2026},
  url = {https://doris.apache.org/docs/4.x/sql-manual/sql-functions/ai-functions/ai-filter/}
}

@article{fagin2003optimal,
  title = {Optimal Aggregation Algorithms for Middleware},
  author = {Fagin, Ronald and Lotem, Amnon and Naor, Moni},
  journal = {Journal of Computer and System Sciences},
  volume = {66},
  number = {4},
  pages = {614--656},
  year = {2003},
  doi = {10.1016/S0022-0000(03)00026-6},
  url = {https://doi.org/10.1016/S0022-0000(03)00026-6}
}

@article{hoeffding1963probability,
  title = {Probability Inequalities for Sums of Bounded Random Variables},
  author = {Hoeffding, Wassily},
  journal = {Journal of the American Statistical Association},
  volume = {58},
  number = {301},
  pages = {13--30},
  year = {1963},
  doi = {10.1080/01621459.1963.10500830},
  url = {https://doi.org/10.1080/01621459.1963.10500830}
}

@article{lao2026sembench,
  title = {{SemBench}: A Benchmark for Semantic Query Processing Engines},
  author = {Lao, Jiale and Zimmerer, Andreas and Ovcharenko, Olga and Cong, Tianji and Russo, Matthew and Vitagliano, Gerardo and Cochez, Michael and {\"O}zcan, Fatma and Gupta, Gautam and Hottelier, Thibaud and Jagadish, H. V. and Kissel, Kris and Schelter, Sebastian and Kipf, Andreas and Trummer, Immanuel},
  journal = {Proceedings of the VLDB Endowment},
  volume = {19},
  number = {8},
  pages = {1754--1767},
  year = {2026},
  doi = {10.14778/3811243.3811249},
  url = {https://www.vldb.org/pvldb/vol19/p1754-trummer.pdf}
}

@article{lee2025saber,
  title = {{SABER}: A {SQL}-Compatible Semantic Document Processing System Based on Extended Relational Algebra},
  author = {Lee, Changjae and Zhao, Zhuoyue and Xiong, Jinjun},
  journal = {arXiv preprint arXiv:2509.00277},
  year = {2025},
  doi = {10.48550/arXiv.2509.00277},
  url = {https://arxiv.org/abs/2509.00277}
}

@article{lin2024zendb,
  title = {Towards Accurate and Efficient Document Analytics with Large Language Models},
  author = {Lin, Yiming and Hulsebos, Madelon and Ma, Ruiying and Shankar, Shreya and Zeigham, Sepanta and Parameswaran, Aditya G. and Wu, Eugene},
  journal = {arXiv preprint arXiv:2405.04674},
  year = {2024},
  doi = {10.48550/arXiv.2405.04674},
  url = {https://arxiv.org/abs/2405.04674}
}

@inproceedings{liu2024suql,
  title = {{SUQL}: Conversational Search over Structured and Unstructured Data with Large Language Models},
  author = {Liu, Shicheng and Xu, Jialiang and Tjangnaka, Wesley and Semnani, Sina J. and Yu, Chen Jie and Lam, Monica S.},
  booktitle = {Findings of the Association for Computational Linguistics: NAACL 2024},
  pages = {4535--4555},
  year = {2024},
  doi = {10.18653/v1/2024.findings-naacl.283},
  url = {https://doi.org/10.18653/v1/2024.findings-naacl.283}
}

@inproceedings{liu2025palimpzest,
  title = {{Palimpzest}: Optimizing {AI}-Powered Analytics with Declarative Query Processing},
  author = {Liu, Chunwei and Russo, Matthew and Cafarella, Michael J. and Cao, Lei and Chen, Peter Baile and Chen, Zui and Franklin, Michael J. and Kraska, Tim and Madden, Samuel and Shahout, Rana and Vitagliano, Gerardo},
  booktitle = {Proceedings of the Conference on Innovative Data Systems Research (CIDR)},
  year = {2025}
}

@book{manning1999foundations,
  title = {Foundations of Statistical Natural Language Processing},
  author = {Manning, Christopher D. and Sch{\"u}tze, Hinrich},
  publisher = {MIT Press},
  address = {Cambridge, MA},
  year = {1999}
}

@article{qi2026sema,
  title = {{Sema}: A High-performance System for {LLM}-based Semantic Query Processing},
  author = {Qi, Kangkang and Xie, Dongyang and Li, Wenbo and Zhang, Hao and Zhu, Yuanyuan and Yu, Jeffrey Xu and Zhao, Kangfei},
  journal = {arXiv preprint arXiv:2603.11622},
  year = {2026},
  doi = {10.48550/arXiv.2603.11622},
  url = {https://arxiv.org/abs/2603.11622}
}

@inproceedings{schick2023toolformer,
  title = {{Toolformer}: Language Models Can Teach Themselves to Use Tools},
  author = {Schick, Timo and Dwivedi-Yu, Jane and Dess{\`i}, Roberto and Raileanu, Roberta and Lomeli, Maria and Hambro, Eric and Zettlemoyer, Luke and Cancedda, Nicola and Scialom, Thomas},
  booktitle = {Advances in Neural Information Processing Systems},
  volume = {36},
  pages = {68539--68551},
  year = {2023}
}

@article{shankar2025docetl,
  title = {{DocETL}: Agentic Query Rewriting and Evaluation for Complex Document Processing},
  author = {Shankar, Shreya and Chambers, Tristan and Shah, Tarak and Parameswaran, Aditya G. and Wu, Eugene},
  journal = {Proceedings of the VLDB Endowment},
  volume = {18},
  number = {9},
  pages = {3035--3048},
  year = {2025},
  doi = {10.14778/3746405.3746426},
  url = {https://doi.org/10.14778/3746405.3746426}
}

@misc{snowflake2026aisql,
  title = {Snowflake Cortex {AI} Functions},
  author = {{Snowflake}},
  howpublished = {Snowflake Documentation},
  year = {2026},
  note = {Retrieved June 9, 2026},
  url = {https://docs.snowflake.com/en/user-guide/snowflake-cortex/aisql}
}

@misc{snowflake2026aifilter,
  title = {{AI\_FILTER}},
  author = {{Snowflake}},
  howpublished = {Snowflake Documentation},
  year = {2026},
  note = {Retrieved July 29, 2026},
  url = {https://docs.snowflake.com/en/sql-reference/functions/ai_filter}
}

@article{tang2025dataanalystsurvey,
  title = {{LLM}/Agent-as-Data-Analyst: A Survey},
  author = {Tang, Zirui and Wang, Weizheng and Zhou, Zihan and Jiao, Yang and Xu, Bangrui and Niu, Boyu and Zhou, Xuanhe and Li, Guoliang and He, Yeye and Zhou, Wei and Song, Yitong and Tan, Cheng and Wang, Bin and He, Conghui and Wang, Xiaoyang and Wu, Fan},
  journal = {arXiv preprint arXiv:2509.23988},
  year = {2025},
  doi = {10.48550/arXiv.2509.23988},
  url = {https://arxiv.org/abs/2509.23988}
}

@inproceedings{theobald2004topk,
  title = {Top-k Query Evaluation with Probabilistic Guarantees},
  author = {Theobald, Martin and Weikum, Gerhard and Schenkel, Ralf},
  booktitle = {Proceedings of the 30th International Conference on Very Large Data Bases (VLDB)},
  pages = {648--659},
  year = {2004}
}

@misc{anthropic2025advancedtooluse,
  title = {Introducing Advanced Tool Use on the {Claude} Developer Platform},
  author = {{Anthropic}},
  year = {2025},
  url = {https://www.anthropic.com/engineering/advanced-tool-use}
}
\bibliographystyle{icml2026}

\newpage
\appendix
\onecolumn
\providecommand{\tabench}{TABench}

\section{Additional Discussion and Scope}
\label{app:discussion}

\subsection{What is atomic in TADA?}

Tagging and Categorization are atomic at the interface boundary: each exposes
one stable output structure to the analytical workflow. Atomicity does not
mean that an implementation uses one prompt, one LLM call, or no internal
symbolic processing. A Tagging call may perform intent detection, batching,
validation, and recovery before returning one string per input item.
Categorization may generate local candidates and consolidate them before
returning one category list. The caller observes the same contract regardless
of these internal choices.

The atom-level query specifies the semantic processing to perform. It may
include a perspective, allowed labels, a scale, or a category count, but it
does not redefine the output structure. Tagging always returns one string per
input item; Categorization always returns category objects with a required
name and optional description. This separation lets the semantic task vary
without expanding the public API.

\subsection{Strings, types, and symbolic validation}

Tagging uses strings as a common interchange representation. Depending on the
query, a returned string may denote a Boolean decision, category name, number,
extracted span, translation, or summary. The symbolic runtime is responsible
for converting values to the required type and validating task-specific
constraints before using them in filtering, sorting, aggregation, or model
fitting. The contract therefore provides representational flexibility but
does not by itself guarantee that every returned string is valid for its
downstream operation.

\subsection{Relationship to semantic operators}

Many row-level semantic operators can be decomposed into Tagging followed by
an exact operation, and category-discovery workflows can be expressed through
Categorization followed, when needed, by Tagging. The decomposition does not
make the atoms more expressive than an operator system with a semantic map;
the same workflows can be written there. It is most
useful when later steps need the semantic state itself. A composite operator
remains preferable when the final filtered, ranked, joined, or aggregated
artifact is the only required result, or when its implementation uses
specialized pairwise or retrieval algorithms.

The two atoms are not a completeness claim. Pairwise joins,
comparison-based ranking, graph construction, global matching, and other
structured outputs may motivate additional atoms. The criterion is a new
semantic output structure that cannot be represented through the existing
contracts, rather than a new sequence of symbolic operations over outputs the
two atoms already expose.

\subsection{What scale transparency guarantees}

Scale transparency means that the public signature and output structure do
not change with input size. It does not assert that a one-call and multi-call
execution produce identical semantic values. Partitioning changes the context
seen by local calls, candidate retention may discard information, and
semantic consolidation may merge or split categories differently. Validation,
retries, completion checks, and the large-scale experiments in the main paper
characterize these implementation-level risks.

\section{Atom and Agent Implementation Materials}
\label{app:implementation}

\begin{table*}[t]
\centering
\small
\caption{Public atom contracts and their materialized analytical state.}
\label{tab:atom-contract-summary}
\begin{tabular}{p{0.11\textwidth}p{0.21\textwidth}p{0.19\textwidth}p{0.19\textwidth}p{0.18\textwidth}}
\toprule
\textbf{Atom} & \textbf{Input} & \textbf{Raw return} &
\textbf{Materialization} & \textbf{Large-scale path} \\
\midrule
Tagging &
Text sequence $X$ and atom-level query $q$ &
One string $s_i$ for each $x_i$, preserving length and order &
One row-aligned column associated with input \texttt{rid} values &
Independent: partition and reassemble; joint: discover a shared category
list, then use the independent path \\
\addlinespace
Categorization &
Text sequence $X$ and atom-level query $q$ &
Category objects with a required name and optional description &
One category table with fresh \texttt{cid} values &
Partition, retain local candidates, semantically consolidate, and select the
requested global list \\
\bottomrule
\end{tabular}
\end{table*}

\subsection{Prompting and model roles}

The atom library combines a dedicated system prompt, the atom-level query, and
the selected text values. The prompts adapt the Algorithmic Prompting and
Thinking-before-Speaking principles used by CAST to the current Tagging and
Categorization contracts. The orchestrator model is selected independently:
it analyzes the user request, inspects the table schema, writes the complete
Python workflow, and responds to execution failures. Atom models execute the
narrow, high-volume semantic calls. Recording both model identifiers is
therefore necessary to reproduce a TADA-Agent run.

\subsection{Programmatic Tool Calling and the text-analysis skill}

The text-analysis skill supplies the orchestrator with the two API signatures,
input and output conventions, composition rules, and canonical examples. The
orchestrator then invokes \texttt{ta.tag(data, column, query)} and
\texttt{ta.categorize(data, column, query)} inside generated Python. Dataframe
slices, loop variables, atom outputs, and exceptions remain in the same
code-execution environment.

\subsection{Scale-transparent execution and recovery}

For independent Tagging, each chunk is processed under the same query and the
library verifies that every input row has exactly one output before
reassembling by \texttt{rid}. For joint Tagging, large-scale Categorization
first establishes a shared category list and the library constructs one
assignment query for every chunk. For Categorization, the library shuffles
rows, produces local candidate lists under budget $r$, and semantically
consolidates all local lists and pilot support evidence in one global call.
The implementation rejects a reducer input that exceeds its conservative
context budget rather than introducing a multi-stage reducer.

For each reported large-scale run, we record $N$, $k$, $n$, $r$, the realized
chunk sizes, model identifiers, prompts, random seeds, mapper and reducer
calls, retries, completion checks, token usage, and wall time. Missing rows,
duplicate identifiers, malformed category objects, under-filled local lists,
and failed consolidation calls are execution outcomes rather than ignorable
logging events.

\section{Top-\texorpdfstring{$k$}{k} Preservation and Bootstrap Planning}
\label{app:topk-bootstrap}

\subsection{Fixed-taxonomy preservation analysis}

Consider the idealized candidate-retention layer used to guide the
large-scale Categorization planner.  Let
$\mathcal L=(\ell_1,\ldots,\ell_C)$ be fixed canonical categories, and assign
each item exactly one category through a partition-invariant function $Y$.
Order the categories by their full-input support:
\begin{equation}
    p_i
    =
    \frac{1}{N}
    \sum_{x\in X}\mathbf 1[Y(x)=i],
    \qquad
    p_1\geq p_2\geq\cdots\geq p_C.
    \label{app:eq:fixed-label-support}
\end{equation}

After a random shuffle, let $X_1,\ldots,X_B$ be balanced chunks and let
$n_{\min}=\min_b|X_b|$.  Each chunk computes exact category counts and retains its
local top-$r$:
\begin{align}
    A_b(i)
    &=
    \sum_{x\in X_b}\mathbf 1[Y(x)=i],\\
    H_b
    &=
    \textsc{TopR}(A_b,r),\\
    \widehat T_k
    &=
    \textsc{TopK}\!\left(
      \textsc{SumCanonicalCounts}(H_1,\ldots,H_B),
      k
    \right).
    \label{app:eq:fixed-taxonomy-combination}
\end{align}
Here \textsc{TopR} returns the $r$ categories with largest chunk counts
together with those counts, using a fixed tie rule, and
\textsc{SumCanonicalCounts} combines retained counts by immutable category
identifier.  Assume $k\leq r\leq C$, define $p_{C+1}=0$, and let
\begin{equation}
    \Delta_{i,r}=p_i-p_{r+1}.
    \label{app:eq:rank-sensitive-gap}
\end{equation}
Let $T_k=\{1,\ldots,k\}$ denote the full-input top-$k$.

\begin{theorem}[Top-$k$ preservation]
\label{app:thm:topk-preservation}
Assume $p_k>p_{k+1}$; $Y$ is fixed independently of the random partition;
each chunk computes exact counts and exact local top-$r$ categories; and the
reducer sums retained counts by canonical identifier.  Then
\begin{equation}
    \Pr\!\left(\widehat T_k=T_k\right)
    \geq
    1-
    B(C-r)
    \sum_{i=1}^{k}
    \exp\!\left(
      -\frac{n_{\min}\Delta_{i,r}^{2}}{2}
    \right),
    \label{app:eq:topk-preservation-bound}
\end{equation}
where the probability is over the random partition, conditional on
$\mathcal L$ and $Y$.
\end{theorem}

\paragraph{Proof.}
Let
\(
\widehat S(i)=\sum_{b=1}^{B}A_b(i)\mathbf 1[i\in H_b]
\)
be the count observed by the reducer.  Suppose no positive count of a
full-input top-$k$ category is discarded.  The chunks partition $X$, so
\(\widehat S(i)=Np_i\) for every $i\leq k$, whereas
\(\widehat S(j)\leq Np_j\) for every $j>k$.  Hence
\[
    \widehat S(i)
    \geq Np_k
    > Np_{k+1}
    \geq Np_j
    \geq \widehat S(j)
    \qquad (i\leq k<j),
\]
and the reducer returns $T_k$.

If $A_b(i)>0$ for some $i\leq k$ but $i\notin H_b$, then at least one
retained category $j>r$ ties or overtakes it.  For a random chunk, the
difference
\(\mathbf 1[Y(x)=i]-\mathbf 1[Y(x)=j]\)
lies in $[-1,1]$ and has mean at least $\Delta_{i,r}$.
Hoeffding's inequality for sampling without replacement bounds this inversion
by
\(\exp(-|X_b|\Delta_{i,r}^2/2)\)
~\citep{hoeffding1963probability}.
A union bound over $B$ chunks, $k$ protected categories, and $C-r$ tail
categories yields Equation~\ref{app:eq:topk-preservation-bound}.  Ties are
included in the failure event.  When $r=C$, no category is discarded and
recovery is exact.

For $r<C$, the smallest protected gap gives the sufficient condition
\begin{equation}
    n_{\min}
    \geq
    \frac{2}{\Delta_{k,r}^{2}}
    \ln\!\left(
      \frac{Bk(C-r)}{\delta}
    \right)
    \quad\Longrightarrow\quad
    \Pr(\widehat T_k=T_k)\geq1-\delta.
    \label{app:eq:chunk-size-rule}
\end{equation}
The condition uses the realized minimum chunk size
$n_{\min}=\lfloor N/B\rfloor$ with $B=\lceil N/n\rceil$.

\subsection{Theory-guided operational bootstrap}
\label{app:bootstrap-planner}

The sufficient bound explains which quantities govern retention but can be
vacuous and is sensitive to plugging noisy pilot frequencies into a squared
gap.  TADA therefore records the plug-in bound as a diagnostic and uses a
paired bootstrap to select $r$.

Let the pilot produce canonical assignments
$S^{\mathrm p}=(s_1^{\mathrm p},\ldots,s_{N_{\mathrm p}}^{\mathrm p})$.
For each outer bootstrap replicate $u$, the planner:
\begin{enumerate}
    \item draws a support profile from the Jeffreys Dirichlet posterior
          $\operatorname{Dirichlet}(c_1+\tfrac12,\ldots,c_C+\tfrac12)$,
          where $c_i$ is the pilot count of category $i$;
    \item draws multiple random balanced-partition count tables under the
          resampled profile;
    \item for each partition simulation $v$, computes the realized full-input
          top-$k$, denoted $T_k^{(u,v)}$;
    \item for every candidate $r\in\mathcal R$, retains local top-$r$ counts,
          combines them by canonical identifier, and records whether
          $\widehat T_k^{(u,v)}=T_k^{(u,v)}$.
\end{enumerate}
All candidate values share the same support profiles and partition draws.  If
$Z_{u,v}(r)\in\{0,1\}$ is the recovery indicator, the conditional recovery
for outer replicate $u$ is
\[
    \widehat\rho_u(r)
    =
    \frac{1}{V}\sum_{v=1}^{V}Z_{u,v}(r).
\]
The implementation takes a lower empirical quantile $Q_{\alpha}(r)$ across
$\widehat\rho_1(r),\ldots,\widehat\rho_U(r)$ and selects the smallest
candidate whose quantile meets the target recovery.  This quantile represents
pilot-profile uncertainty; the finite number of partition simulations is
recorded separately as Monte Carlo error rather than being presented as a
deployment guarantee.  This makes the theoretical connection explicit
without treating either the conservative sufficient bound or simulation error
as a calibrated probability estimate.

The planner records, for every candidate $r$, the empirical gap
$\widehat p_k-\widehat p_{r+1}$, the plug-in theorem lower bound, bootstrap
mean recovery, lower pilot-profile quantile, Monte Carlo diagnostics, and
candidate-retention volume $Br$.  If the
pilot exposes no category below rank $k$, $r$ is not identifiable from the
observed tail; the implementation uses a conservative configured budget and
records that fallback.

\subsection{Scope of the analysis}

The theorem and bootstrap isolate candidate retention under one canonical
label per row and exact consolidation.  Live Categorization additionally
contains semantic discovery, local under-filled lists, synonymous category
names, and consolidation error.  The experiments therefore report an
evaluation with known full-population labels, a fixed semantic-assignment
layer, and a separate
end-to-end open-discovery layer rather than attributing every failure to
$n$ or $r$.  The bootstrap also treats the observed pilot category list as
the available tail and the pilot Tagging outputs as canonical.  Unseen
categories, pilot-to-full taxonomy drift, Tagging error, and the difference
between count-ranked local top-$r$ and LLM-ranked category proposals are
reported as limitations rather than absorbed into the retention estimate.

\section{Neural--Symbolic Boundary Diagnostic}
\label{app:boundary-diagnostic}

We use a controlled diagnostic to test the execution
boundary motivating TADA. Each task receives the full synthetic table in one
GPT-4.1 call. The \emph{monolithic} condition asks the model to perform semantic
processing together with sorting, filtering, or aggregation. The
\emph{atom+symbolic} condition uses Tagging to extract the row-level semantic values
and Python to perform the exact operation. No agent orchestration, chunking,
consolidation, or recovery is used.

\begin{table*}[t]
\centering
\small
\caption{Single-run neural--symbolic boundary diagnostic. Sort
coverage reports returned identifiers over requested rows; aggregation reports
exact agreement on count, sum, and average.}
\label{tab:boundary-diagnostic}
\begin{tabular}{rrrrc}
\toprule
$N$ & \multicolumn{1}{c}{Monolithic sort} &
\multicolumn{1}{c}{Monolithic filter F1} &
\multicolumn{1}{c}{Monolithic agg.} &
\multicolumn{1}{c}{Atom+symbolic} \\
\midrule
100  & 79/100   & 1.000 & No & Exact \\
200  & 161/200  & 0.765 & No & Exact \\
500  & 280/500  & 0.678 & No & Exact \\
1000 & 400/1000 & 0.105 & No & Exact \\
\bottomrule
\end{tabular}
\end{table*}

Tagging followed by symbolic code is exact for all three tasks at every tested
size. The monolithic sort is incomplete at every size, the filter degrades
sharply as $N$ grows, and every monolithic aggregate is numerically incorrect.
At $N=1000$, the sort output also contains three duplicate identifiers.

This diagnostic isolates the neural--symbolic boundary on templated inputs with
rule-based answers. Each scale point has one GPT-4.1 execution, and the
monolithic condition uses a raw prompt rather than a mature semantic-operator
system; it therefore illustrates a mechanism rather than estimating comparative
system performance.

\section{Public-Benchmark Protocol Details}
\label{app:public-benchmarks}

\subsection{CUAD}
\label{app:cuad}

CUAD is implemented as row-aligned clause Tagging followed by deterministic
normalization into the official evaluation format. The main text reports
0.787 average F1 for TADA-Agent versus 0.753 for DocETL and 0.619 for
LOTUS under the common evaluation environment.

\subsection{BioDEX}
\label{app:biodex}

BioDEX contains 100 drug-safety reports evaluated against the 24,313-term
MedDRA vocabulary. TADA first extracts free-text adverse-reaction phrases
through Tagging and then deterministically aligns them to the top-200 retrieved
candidate terms. Three saved TADA runs yield mean (SD) values of 0.2295
(0.0009) RP@5, 0.2465 (0.0095) RP@10, and 0.3849 (0.0104) term recall. Their
95\% $t$ intervals are [0.2273, 0.2317], [0.2229, 0.2700], and [0.3591,
0.4107], respectively. The protocol-matched DocETL point is
(0.201, 0.196, 0.305) and the LOTUS point is (0.194, 0.200, 0.334), ordered as
RP@5, RP@10, and term recall. Other saved LOTUS runs use different prompting
and retrieval settings and are excluded from the matched comparison.

\section{TABench Construction, Task Suite, and Evaluation}
\label{app:tabench-construction}
\label{app:tabench}
\label{app:tabench-protocol}

\subsection{Taxonomy-derived construction procedure}

TABench was constructed from a literature-grounded taxonomy of user intents over
free-text columns. The taxonomy has nine top-level classes: classification and
labeling, open grouping and clustering, semantic filtering, information
extraction and text structuring, aggregation and summarization, joint analysis
of text and structured columns, semantic ranking, anomaly detection and data
repair, and semantic augmentation. For benchmark annotation, three broad
classes are split where their output contracts differ materially: field
extraction versus text-to-table structuring, numerical/temporal analysis
versus semantic join, and anomaly detection versus repair. This yields the 12
scenario classes used in the main paper (S1--S12 in
Table~\ref{tab:tabench-inventory-classes}): S1~row-wise classification,
S2~open grouping and clustering, S3~semantic filtering, S4~field extraction,
S5~text-to-table structuring, S6~aggregation and summarization,
S7~numerical/temporal analysis, S8~semantic join, S9~semantic ranking,
S10~anomaly detection, S11~normalization and repair, and S12~constrained
generation and augmentation.

Case construction followed four stages. First, each scenario class was mapped to
at least one realistic workbook intent and to the semantic and deterministic
operations needed to answer it. Second, source workbooks were generated with
stable row keys and with enough structured context to exercise downstream
filtering, grouping, joining, aggregation, or ranking rather than only an
isolated LLM call. Third, each request received an explicit output contract
and query-specific rubric covering schema, row or group coverage, deterministic
computation, semantic content, and usability. Fourth, cases were annotated along
analytical scenario, semantic operation, and input--output structure.

The suite was built in two rounds. A seed round produced 26 cases, one per
distinct taxonomy leaf sub-scenario then covered. An expansion round added
17 cases targeting the remaining independent leaf sub-scenarios, in two
groups: new queries over already-frozen source workbooks (each on a distinct
workbook, with answers not derivable from the original task's output), and
newly generated workbooks whose generator injects controlled phenomena
(near-duplicate paraphrase clusters, entity-alias variants, format-diverse
numeric and date fields, misleading auxiliary columns) and therefore carries
deterministic gold labels alongside the judge protocol. Three seed cases
probing text-to-table structuring, semantic imputation, and semantic join
were excluded from the evaluated suite because their required output
structure falls outside the two-atom contract; they are discussed as
boundary evidence in the main paper's Discussion section rather than scored.
The resulting evaluated suite contains 40 cases. It is a coverage-driven
diagnostic sample, not a claim about the empirical frequency of all
real-world spreadsheet requests.

\subsection{Task inventory}

The TABench suite contains 43 source-workbook and
natural-language-request pairs. The scored evaluation uses 40 of them,
organized by the scenario classes of the taxonomy
(Table~\ref{tab:tabench-inventory-classes}). Ten of the 12 classes are
represented; text-to-table structuring (S5) and semantic join (S8) are the
boundary classes excluded above; the remaining excluded request probes
semantic imputation, another output contract not represented by the current
atoms. Every covered class contains at
least three cases; several requests additionally stress multiple columns,
cross-row context, or misleading auxiliary signals. The suite therefore
evaluates complete semantic--symbolic workflows rather than isolated
calls to a preselected semantic operator.

\begin{table*}[t]
\centering
\small
\setlength{\tabcolsep}{4pt}
\caption{TABench task inventory by scenario class. Dominant program
structure describes the typical composition; a query may exercise more
than one class's mechanics. S5 and S8 are boundary classes outside the
two-atom contract and are discussed rather than scored.}
\label{tab:tabench-inventory-classes}
\begin{tabular}{llp{0.16\textwidth}p{0.40\textwidth}}
\toprule
\textbf{ID} & \textbf{Scenario class} & \textbf{Cases} & \textbf{Dominant program structure} \\
\midrule
S1 & Row-wise classification & Q01--Q03, Q27, Q28 & Tag typed labels against a fixed or inferred label set (one string value per row per call). \\
S2 & Open grouping \& clustering & Q04, Q05, Q36 & Categorize a corpus-level taxonomy, optionally Tag rows with its identifiers; variants add outlier isolation. \\
S3 & Semantic filtering & Q06--Q08, Q29, Q38 & Tag a predicate per row, then deterministic row selection; includes negated, implicit, and alias-sensitive predicates. \\
S4 & Field extraction & Q09, Q10, Q30, Q39, Q40 & Tag typed fields per row from free text; repeated per output column; includes open-set and relational attributions. \\
S6 & Aggregation \& summarization & Q13--Q15, Q31 & Group rows deterministically, Categorize or summarize per group, compare across groups or time. \\
S7 & Numerical/temporal analysis & Q16, Q17, Q32, Q42 & Combine semantic state with numeric filtering, aggregation, or multi-column fusion. \\
S9 & Semantic ranking & Q19, Q20, Q33, Q34 & Tag scores on the requested numerical scale or ordering constructs, then deterministic sort. \\
S10 & Anomaly detection & Q21, Q22, Q37 & Categorize or Tag to expose outliers, near-duplicates, or emerging segments. \\
S11 & Normalization \& repair & Q23, Q44, Q45 & Tag canonical values or suspect flags preserving the original column. \\
S12 & Constrained generation \& augmentation & Q24--Q26, Q35 & Tag generated columns under explicit output constraints. \\
\bottomrule
\end{tabular}
\end{table*}

\subsection{Contract checks and rule-based answers}

The normalization-and-repair class (S11) is reported here because its row-level
metric evaluates a single case; the remaining cases use rule-based contract
checks. On the row-level case all five methods lie between 99.8 and 100.0 with
fully overlapping intervals. On the seven cases with rule-based answers,
TADA-Agent scores
exactly 100\% on all of them; the semantic-operator baseline is at 100\%
on five, with 99.9$\pm$0.1 and 98.2$\pm$3.2 on the remaining two.

\subsection{Semantic-operator execution and request ledger}

The semantic-operator baseline runs through the library's official
\texttt{sem\_*} APIs with unchanged prompts, model, temperature, and batch
size on the same fixed endpoint. Two engineering accommodations were
required and are disclosed: the library's \texttt{sem\_extract} JSON mode
is incompatible with the serving endpoint's structured-output interface,
so affected cases use a semantically equivalent per-field \texttt{sem\_map}
fallback; and one long-running case was completed through 25-row
checkpointed chunks after a host interruption, with all rows preserved.
The request ledger classifies every task-run's evidence as exact
(persisted per-call progress), reconstructed (deterministic per-row
arithmetic from completed artifacts), or bounded (randomized traces with
provable ranges). Judge calls and host-agent turns are excluded. The
reported total of at least 18{,}852 requests is a strict lower bound: one
failed task-run defines 400 planned per-row requests but preserves no
durable progress and no deliverable, so it contributes zero; failed and
recovery attempts without persistent usage are likewise excluded. Full
token totals are not reported because per-request usage was not
persistently recorded by the library, and wall time is not used as
evidence because endpoint and substrate latency confound it.

\subsection{Judge protocol and calibration}

The unified protocol records per-case judgment-unit counts, disputed-unit
fractions, per-run net verdicts, the adjacent-level tie rule for fixed integer
rating scales, and the stratified human-calibration design. The controlled matrix comprises
17{,}949 row-aligned judged units (107{,}694 judge calls) and 266
case-level units (1{,}596 calls); the semantic-operator comparison adds
further row-aligned and case-level units judged under the identical
protocol.
\paragraph{Stratified human calibration.}
The calibration sample contains 39 judgment units drawn from the
row-aligned tasks: 21 from contested units (where per-run judge
verdicts disagreed or direction majorities conflicted), 12 from
low-confidence units, and 6 from high-confidence units. Six human
annotators independently and blindly labeled every unit under the same
anonymized A/B presentation as the judge (A better / B better / tie),
without access to system identities or judge verdicts. Annotators reach
a majority verdict on 33 of 39 units; the judge matches the human
majority on 26 of these 33 (78.8\%, Cohen's $\kappa=0.68$), with two
direction reversals and the remainder tie-versus-preference
disagreements. Table~\ref{tab:human-calibration} breaks the comparison
down by stratum and task. Inter-annotator agreement is itself moderate
(mean pairwise raw agreement 0.55, Fleiss $\kappa=0.32$), and the
judge's mean raw agreement with individual annotators (0.60) falls
within the annotator pool's own range (0.36--0.77). All six
majority-free units fall on fixed-scale rating tasks where the candidates
differ by one scale step, the regime absorbed by the deterministic
tie-band rule.
The sample was drawn from an earlier calibration pool and includes Q12, which
is used only for judge calibration and does not enter the final 40-case system
scores.

\begin{table}[t]
\caption{Human calibration of the LLM judge on 39 stratified judgment
units, six annotators. \emph{Maj.} = units with a human majority
verdict; \emph{Agree} = judge matches human majority; \emph{Flip} =
directional conflict (one side A, the other B).}
\label{tab:human-calibration}
\centering
\footnotesize
\begin{tabular}{lcccc}
\toprule
 & \textbf{Units} & \textbf{Maj.} & \textbf{Agree} & \textbf{Flip} \\
\midrule
\multicolumn{5}{l}{\emph{By stratum}} \\
Contested       & 21 & 19 & 17 & 1 \\
Low-confidence  & 12 &  8 &  4 & 1 \\
High-confidence &  6 &  6 &  5 & 0 \\
\midrule
\multicolumn{5}{l}{\emph{By task type}} \\
Summarization (Q24)          & 8 & 8 & 8 & 0 \\
Translation (Q25)            & 7 & 7 & 7 & 0 \\
Document revision (Q12)      & 6 & 5 & 1 & 1 \\
Relevance rating (Q20)       & 6 & 4 & 2 & 1 \\
Multi-column extraction (Q10) & 4 & 4 & 4 & 0 \\
Priority rating (Q26)        & 4 & 1 & 1 & 0 \\
Sentiment rating (Q19)       & 3 & 3 & 2 & 0 \\
Urgency triage (Q02)         & 1 & 1 & 1 & 0 \\
\midrule
Total & 39 & 33 & 26 & 2 \\
\bottomrule
\end{tabular}
\end{table}

\subsection{Error-analysis procedure}

After scoring, evaluator notes and available traces
were inspected to classify recurring failure modes. These include
lexical or rule-based substitution for a semantic judgment, incoherent
open-ended categories, incomplete row coverage, incorrect symbolic
post-processing, missing workbook mutations, and tool failures. These
categories are diagnostic rather than additional scoring rules.

\section{Large-Scale End-to-End Execution Details}
\label{app:e2e-details}

This appendix documents per-repetition execution and cost telemetry for
the RQ2 end-to-end comparison in the main paper. All rows
correspond to recorded runs; nothing is estimated except where
explicitly marked.

\subsection{Per-repetition cost telemetry}
\label{app:e2e-cost}

Table~\ref{tab:e2e-cost} lists every repetition. The three systems are
metered differently, and the table preserves those distinctions rather
than merging them: TADA-Agent and LOTUS token counts are provider-reported
physical usage; Claude Code counts are recovered from its session
transcripts (one entry per assistant API message), where token totals are
dominated by prompt-cache reads and three repetitions have incomplete
token fields (marked $\ast$), which we leave unavailable rather than
estimate.

\begin{table}[h]
\caption{Per-repetition execution cost. Calls: planned LLM calls for
TADA-Agent (observed attempts in parentheses where they differ), API
messages for Claude Code, per-row calls for LOTUS. Tokens in thousands;
``cache'' is Claude Code prompt-cache reads. Wall time in minutes.
$\ast$: transcript token fields incomplete; --: not recorded or not
aggregable.}
\label{tab:e2e-cost}
\centering
\footnotesize
\setlength{\tabcolsep}{4pt}
\begin{tabular}{llrrrrr}
\toprule
Workload & Rep & Calls & Prompt & Compl. & Cache & Wall \\
\midrule
\multicolumn{7}{l}{\emph{TADA-Agent (provider-reported)}}\\
Sentiment & 1 & 33 (105) & -- & -- & -- & 164.6 \\
Sentiment & 2 & 33 & 891.7 & 349.2 & -- & 15.1 \\
Sentiment & 3 & 33 & 891.7 & 259.9 & -- & 12.2 \\
Topic & 1 & 33 & 894.7 & 623.5 & -- & 29.2 \\
Topic & 2 & 33 & 894.7 & 674.5 & -- & 31.8 \\
Topic & 3 & 33 & 894.7 & 656.0 & -- & 36.8 \\
Category & 1 & 7 & 242.5 & 49.5 & -- & 2.3 \\
Category & 2 & 7 & 242.5 & 56.5 & -- & 9.6 \\
Category & 3 & 7 & 242.5 & 95.4 & -- & 5.4 \\
\midrule
\multicolumn{7}{l}{\emph{Claude Code (transcript-recovered)}}\\
Sentiment & 1 & 67 & 93.1 & 51.0 & 39.4 & 9.1 \\
Sentiment & 2 & 99 & 131.6 & 65.1 & 734.6 & 10.0 \\
Sentiment & 3 & 40 & 5.5$^\ast$ & 20.7 & 12.0$^\ast$ & 29.4 \\
Topic & 1 & 40 & 57.9 & 35.0 & 20.6 & 7.9 \\
Topic & 2 & 106 & 71.2 & 27.6 & 441.4 & 106.2 \\
Topic & 3 & 155 & 260.2 & 49.9 & 850.5 & 67.9 \\
Category & 1 & 22 & 16.9 & 12.3 & 21.3 & 2.0 \\
Category & 2 & 57 & 39.4 & 23.2 & 142.8 & 3.9 \\
Category & 3 & 29 & --$^\ast$ & 15.3 & --$^\ast$ & 2.4 \\
\midrule
\multicolumn{7}{l}{\emph{LOTUS \texttt{sem\_map} (provider-reported)}}\\
Sentiment & s500 & 500 & 59.8 & 2.5 & -- & 39.3 \\
Topic & s500 & 500 & 91.8 & 2.8 & -- & 33.7 \\
Category & p2050 & 2{,}050 & -- & -- & -- & -- \\
\bottomrule
\end{tabular}
\end{table}

Three notes qualify the table. First, TADA-Agent sentiment repetition~1
is the telemetry exception disclosed in the main text: transient provider
failures inflated the planned 33 calls to 105 observed request events, the
wall time to 164.6 minutes, and the provider token report was
lost. The chunking plan was unchanged; repetitions 2--3 executed it
exactly. Second, the Claude Code topic repetition shown as repetition~2
replaced an earlier attempt that exceeded a three-hour timeout without
producing output. Third, the LOTUS category row aggregates the first
2{,}050 frame rows of a checkpointed full-frame run that spanned several
resumed processes at varying concurrency; its calls are exact but wall
time and tokens are not aggregable across processes and are therefore
omitted.

\subsection{Full-scale cost estimates for the per-row baseline}
\label{app:e2e-lotus-estimate}

LOTUS \texttt{sem\_map} issues one LLM call per input row, so per-row cost
is additive and the measured subsets scale linearly. From the 500-row
measurements (119.6/183.6 prompt and 5.0/5.5 completion tokens per row;
4.7/4.0 seconds per row at concurrency one), one full 32{,}179-row
repetition is estimated at 32{,}179 calls (exact by definition;
975$\times$ TADA-Agent's 33), 4.0M tokens for sentiment and 6.1M for topic
(3--5$\times$ TADA-Agent's measured 1.2--1.6M), and roughly 36--42 hours
of wall time at the same serving concurrency ($>$100$\times$). Call count
is exact; token and time figures are linear extrapolations of measured
per-row cost, not measurements.

\subsection{Coding-agent repetition artifacts}
\label{app:e2e-strategies}

Each controlled coding-agent repetition leaves a saved output file, an exit status,
and a session transcript. Transcript-recovered telemetry
(Table~\ref{tab:e2e-cost}) shows the agent's self-chosen execution
structure varies widely across completed repetitions of the same task: 22--155
assistant API messages, with heavier repetitions corresponding to
iterative script debugging and re-reads of intermediate files (visible as
cache-read-dominated token totals). One topic repetition exceeded a
three-hour budget without producing output and was replaced by a
completed rerun under the identical protocol; the replacement is the
repetition reported in the main table.

\subsection{Open-vocabulary mapping protocol}
\label{app:e2e-openvocab}

In the open-vocabulary variant, the coding agent must first discover its
own category vocabulary. Because discovered labels need not match
reference names, we score them through an explicit semantic label
mapping: each discovered label is mapped to the closest reference
category, or to none when no reference category corresponds; rows whose
labels map to none are counted as errors against the full denominator.
Table~\ref{tab:e2e-openvocab} lists every repetition. Unmapped labels are
of two kinds: granularity merges (one discovered label spanning several
reference categories, e.g.\ ``bugs, crashes \& performance'') and
invented catch-all buckets (``general criticism \& complaints''). 

\begin{table}[h]
\caption{Open-vocabulary repetitions: discovered vocabulary size, labels
with no reference correspondence, and mapped accuracy.}
\label{tab:e2e-openvocab}
\centering
\footnotesize
\begin{tabular}{lrrr}
\toprule
Workload / rep & Discovered & Unmapped & Mapped acc. \\
\midrule
Topic 32K / 1 & 10 & 4 & 0.690 \\
Topic 32K / 2 & 10 & 3 & 0.675 \\
Topic 32K / 3 & 10 & 7 & 0.685 \\
Category 6.6K / 1 & 12 & 8 & 0.296 \\
Category 6.6K / 2 & 11 & 6 & 0.134 \\
Category 6.6K / 3 & 10 & 4 & 0.195 \\
\bottomrule
\end{tabular}
\end{table}

\section{Adaptive Retention Evaluation Details}
\label{app:large-scale-details}
\label{app:rq2-details}

\subsection{Controlled distributions}
\label{app:rq2-controlled}

The controlled retention experiment isolates the fixed-taxonomy retention
stage analyzed in
Appendix~\ref{app:topk-bootstrap}.  It uses $N=32{,}000$ fixed canonical
assignments over $C=20$ categories, requests the global top-$5$, and partitions
the rows into $B=32$ chunks of $n=1{,}000$.  No LLM call occurs during the
retention replay.  This is deliberately separate from open Categorization,
where semantic discovery and consolidation introduce additional errors.

The six populations are generated from explicit descending rank weights.  In
every case,
\[
    (w_1,\ldots,w_5)=(2.00,1.75,1.50,1.25,1.00),
    \qquad
    p_i=\frac{w_i}{\sum_j w_j}.
\]
For the separated cutoff,
\[
    w_6=0.40,
    \qquad
    w_i=0.30(0.85)^{i-7},\quad i=7,\ldots,20.
\]
The five narrow levels share
\[
    w_i=0.75(0.88)^{i-9},\quad i=9,\ldots,20,
\]
and use the following near-cutoff weights:
\[
\begin{array}{c|ccc}
&w_6&w_7&w_8\\\hline
1&0.965&0.955&0.945\\
2&0.970&0.960&0.950\\
3&0.975&0.965&0.955\\
4&0.980&0.970&0.960\\
5&0.982&0.972&0.962
\end{array}
\]
The separated gap is $p_5-p_6=0.061891$; the five narrow gaps are
$0.002293$, $0.001963$, $0.001634$, $0.001306$, and $0.001175$.
We allocate $\lfloor Np_i\rfloor$ rows to each category and distribute the
remaining rows by largest fractional remainder, producing one fixed
32,000-row population per case.

For each population, a 1,000-row sample supplies the planner with canonical
assignments.  The planner draws 100 outer support profiles from a Jeffreys
Dirichlet posterior and evaluates 20 paired random partitions per profile for
every candidate $r\in\{5,\ldots,10\}$.  It chooses the minimum certified
candidate when one exists; otherwise it records a conservative fallback to
the configured maximum $r=10$.

\begin{table*}[t]
\centering
\small
\caption{Planner position relative to the full $r=5,\ldots,20$ recovery
matrix.  The oracle is the first $r$ whose recovery and every larger tested
$r$ remain at least 0.95. Empirical recovery uses 5,000 partitions of each
fixed population.}
\label{tab:rq2-margin-sweep}
\begin{tabular}{lrrrrrr}
\toprule
Regime & $p_5-p_6$ & Oracle $r^\star$ & $\widehat r$ &
$\widehat Q_{.05}$ & Rec.\ at $r=5$ & Rec.\ at $\widehat r$ \\
\midrule
Separated & 0.061891 & 5  & 5  & 1.00 & 1.000 & 1.000 \\
Narrow 1  & 0.002293 & 9  & 10 & 0.90 & 0.909 & 1.000 \\
Narrow 2  & 0.001963 & 9  & 10 & 0.95 & 0.871 & 1.000 \\
Narrow 3  & 0.001634 & 9  & 10 & 0.95 & 0.824 & 1.000 \\
Narrow 4  & 0.001306 & 10 & 10 & 0.95 & 0.744 & 0.990 \\
Narrow 5  & 0.001175 & 10 & 10 & 0.95 & 0.739 & 0.985 \\
\bottomrule
\end{tabular}
\end{table*}

\subsection{Connection to the preservation analysis}
\label{app:rq3-margin}

The theorem and planner use the same quantities: $n_{\min}$, $B$, $k$, $r$,
and the gap from a protected top-$k$ category to rank $r+1$. The plug-in
sufficient bound is zero at both selected operating points, so it is not a
calibrated certificate. It still predicts the qualitative response: larger
cutoff gaps and chunks reduce local rank inversions, while retaining past a
near-tied plateau increases the relevant gap.

The paired bootstrap supplies the operational estimate. Outer profiles
represent pilot-distribution uncertainty; inner simulations estimate
random-partition recovery. Shared draws across $r$ reduce comparison noise.
The separated case needs no extra budget. Across the five narrow levels, the
minimum reliable $r$ is 9 or 10 and the planner recommends 10: exact in the
two hardest levels, conservative by one category in the other three, and never
too small. Recommended-point recovery ranges from 98.5\% to 100\%.

\paragraph{Why not always retain all categories?}
If recovery were the only objective, setting $r=C=20$ would remove
candidate-retention loss by construction.  The planner instead solves a
constrained cost problem: find the smallest local budget that reaches the
target reliability.  Here $B=32$, so $r=10$ sends 320 category objects to the
reducer, whereas $r=20$ sends 640.  The latter also increases mapper output
tokens, reducer context, latency, and---in open Categorization---the number of
low-support or synonymous proposals that semantic consolidation must resolve.
For the hardest narrow level, increasing $r$ from 10 to 20 raises empirical
recovery from 98.5\% to 100\% while doubling candidate volume.  Thus the
selected $r$ is the minimum reliable budget, not the unconstrained
quality-maximizing value.

\paragraph{Shared curve shape and the dip at $r=8$.}
The five orange curves look similar by design: they form a controlled family
that changes only the cutoff gap while keeping the three-category
near-cutoff structure and geometric tail fixed.  Empirical recovery need not be monotone in $r$ because the reducer sums only
retained local counts. Changing $r$ changes
which categories have nearly complete versus systematically truncated
counts.  In narrow level~5, at $r=8$ the average retained-count fractions for
global ranks 5--8 are approximately 96.8\%, 95.6\%, 94.7\%, and 93.9\%.
The three competitors are therefore almost fully represented while the true
rank-5 category still loses enough count for its small global margin to be
reversed.  At $r=9$--10, all four fractions exceed 99\% and recovery rises
sharply.  We therefore define the minimum reliable value as the first
$r$ whose recovery and every larger tested $r$ remain above the target,
rather than the first isolated crossing.

\paragraph{Role of the analysis.}
At these operating points the sufficient bound is vacuous, so the bootstrap is
the selection mechanism. The bound nevertheless identifies the relevant
rank-sensitive margin $\Delta_{k,r}=p_k-p_{r+1}$ rather than only the adjacent
gap $p_k-p_{k+1}$, explaining why retaining past a near-tied plateau can help.
It also exposes the scaling variable $n_{\min}\Delta_{k,r}^2$ and the dependence
on $B$, $k$, and tail size, which determine the bootstrap simulation and the
controlled distributions. It supplies a distribution-free
sufficient certificate in regimes where the bound is nonvacuous and proves
exact recovery at $r=C$, while separating candidate-retention error from
semantic assignment and consolidation error.

\subsection{Scope and evidence}

The controlled experiment characterizes candidate retention under fixed
canonical assignments, complementing the fixed-taxonomy end-to-end Tagging
evaluation in the main paper. Its protocol, population-generating
distributions, and aggregate recovery results are reported here.

\end{document}